\documentclass[review]{elsarticle}

\usepackage{lineno}
\usepackage[cmex10]{amsmath}
\usepackage{booktabs} % For formal tables
\usepackage{epsfig}
\usepackage{graphicx}
\usepackage{multirow}
\usepackage{array}
\usepackage{subfig}
\usepackage{hyperref}
\usepackage{amsmath}
\usepackage{amsthm}
\usepackage{amssymb}
\usepackage{times}
\usepackage{color}
\modulolinenumbers[5]

\journal{Journal of \LaTeX\ Templates}

\begin{document}

\begin{frontmatter}

\title{Deep Time-Frequency Representation and Progressive Decision Fusion for ECG Classification}

%% or include affiliations in footnotes:
\author[1]{Jing Zhang}
\author[1]{Jing Tian}
\author[2]{Yang Cao}
\author[3]{Yuxiang Yang\corref{correspondingauthor}}
\cortext[correspondingauthor]{Corresponding author}
\ead{yyx@hdu.edu.cn}
\author[1]{Xiaobin Xu\corref{correspondingauthor}}
\ead{xuxiaobin1980@163.com}

\address[1]{School of Automation, Hangzhou Dianzi University}
\address[2]{Department of Automation, University of Science and Technology of China}
\address[3]{School of Electronics and Information, Hangzhou Dianzi University}

\begin{abstract}
Early recognition of abnormal rhythms in ECG signals is crucial for monitoring and diagnosing patients' cardiac conditions, increasing the success rate of the treatment. Classifying abnormal rhythms into exact categories is very challenging due to the broad taxonomy of rhythms, noises and lack of large-scale real-world annotated data. Different from previous methods that utilize hand-crafted features or learn features from the original signal domain, we propose a novel ECG classification method by learning deep time-frequency representation and progressive decision fusion at different temporal scales in an end-to-end manner. First, the ECG wave signal is transformed into the time-frequency domain by using the Short-Time Fourier Transform. Next, several scale-specific deep convolutional neural networks are trained on ECG samples of a specific length. Finally, a progressive online decision fusion method is proposed to fuse decisions from the scale-specific models into a more accurate and stable one. Extensive experiments on both synthetic and real-world ECG datasets demonstrate the effectiveness and efficiency of the proposed method.
\end{abstract}


\end{frontmatter}

%\linenumbers

\section{Introduction}
\label{sec:Intro}
Electrocardiogram (ECG), which records the electrical depolarization-repolarization patterns of the heart's electrical activity in the cardiac cycle, is widely used for monitoring or diagnosing patients' cardiac conditions \cite{Giri2013Automated,Acharya2016Automated, kiranyaz2016real}. The diagnosis is usually made by well-trained and experienced cardiologists, which is laborious and expensive. Therefore, automatic monitoring and diagnosing systems are in great demand in clinics, community medical centers, and home health care programs. Although advances have been made in ECG filtering, detection and classification in the past decades \cite{kiranyaz2016real,shyu2004using, guler2005ecg, mar2011optimization}, it is still challenging for efficient and accurate ECG classification due to noises, various types of symptoms, and diversity between patients.

Before classification, a pre-processing filtering step is usually needed to remove a variety of noises from the ECG signal, including the power-line interference, base-line wander, muscle contraction noise, $etc$. Traditional approaches like low-pass filters and filter banks can reduce noise but may also lead some artifacts \cite{wu2009filtering}. Combining signal modeling and filtering together may alleviate this problem, but it is limited to a single type noise \cite{yan2010self, blanco2008ecg}. Recently, different noise removal methods based on wavelet transform has been proposed by leveraging its superiority in multi-resolution signal analysis \cite{bhateja2013composite, jenkal2016efficient, poungponsri2013adaptive}. For instance, S. Poungponsri and X.H. Yu proposed a novel adaptive filtering approach based on wavelet transform and artificial neural networks that can efficiently removal different types of noises \cite{poungponsri2013adaptive}.

For ECG classification, classical methods usually consist of two sequential modules: feature extraction and classifier training. Hand-crafted features are extracted in the time domain or frequency domain, including amplitudes, intervals, and higher-order statistics, $etc$. Various methods have been proposed such as filter banks \cite{afonso1999ecg}, Kalman filter \cite{zeng2016inferring}, Principal Component Analysis (PCA)\cite{Martis2013Characterization, martis2012application}, and wavelet transform (WT) \cite{ince2009generic, jayachandran2010analysis, daamouche2012wavelet, shyu2004using, garcia2016application}. Classifier models including Hidden Markov Models (HMM), Support Vector Machines (SVM) \cite{osowski2004support}, Artificial Neural Networks (ANN) \cite{shyu2004using, guler2005ecg, ince2009generic, barni2011privacy, mar2011optimization, wang2013ecg}, and mixture-of-experts method \cite{hassan2017expert} have also been studied. Among them, a large number of methods are based on artificial neural networks due to its better modeling capacity. For example, L.Y. Shyu $et~al.$ propose a novel method for detecting Ventricular Premature Contraction (VPC) using the wavelet transform and fuzzy neural network \cite{shyu2004using}. By using the wavelet transform for QRS detection and VPC classification, their method has less computational complexity. I. Guler and E.D. Ubeyli propose to use a combined neural network model for ECG beat classification \cite{guler2005ecg}. Statistical features based on discrete wavelet transform are extracted and used as the input of first level networks. Then, sequential networks were trained using the outputs of the previous level networks as input. Unlike previous methods, T. Ince $et~al.$ propose a new method that uses a robust and generic ANN architecture and trains a patient-specific model with morphological wavelet transform features and temporal features for each patient \cite{ince2009generic}. Besides, some approaches have been proposed by combining several hand-crafted features to provide enhanced performance \cite{oster2015impact, li2016signal}. Despite their usefulness, these methods have some common drawbacks: 1) the hand-crafted features rely on domain knowledge of experts and should be designed and tested carefully; 2) the classifier should have appropriate modeling capacity of such features; 2) The types of ECG signals are usually limited.

In the past few years, deep neural networks (DNN) have been widely used in many research fields and achieve remarkable performance. Recently, S. Kiranyaz $et~al.$ propose a 1-D convolutional neural network (CNN) for patient-specific ECG classification \cite{kiranyaz2016real}. They design a simple but effective network architecture and utilize 1-D convolutions to processing the ECG wave signal directly. G. Clifford $et~al.$ organized the PhysioNet/Computing in Cardiology Challenge 2017 for AF rhythm classification from a short single lead ECG recording. A large number of real-world ECG samples from patients are collected and annotated. It facilitates research on the challenging AF classification problem. Both hand-crafted feature-based methods and deep learning-based methods have been proposed \cite{hong2017encase, zabihidetection, teijeiro2017arrhythmia}. For example, S. Hong $et~al.$ propose an ensemble classifier based method by combining expert features and deep features \cite{hong2017encase}. T. Teijeiro $et~al.$ propose a combined two classifiers based method, $i.e.$, the first classifier evaluates the record globally using aggregated values for a set of high-level and clinically meaningful features, and the second classifier utilizes a Recurrent Neural Network fed with the individual features for each detected heartbeat \cite{teijeiro2017arrhythmia}. M. Zabihi $et~al.$ propose a hand-crafted feature extraction and selection method based on a random forest classifier \cite{zabihidetection}.

In this paper, we propose a novel deep CNN based method for ECG classification by learning deep time-frequency representation and progressive decision fusion at different temporal scales in an end-to-end manner. Different from previous methods, 1) We first transform the original ECG signal into the time-frequency domain by Short-Time Fourier Transform (STFT). 2) Then, the time-frequency characteristics at different scales are learned by several scale-specific CNNs with 2-D convolutions. 3) Finally, we propose an online decision fusion method to fuse past and current decisions from different models into a more accurate one. We conducted extensive experiments on a synthetic ECG dataset consisting of 20 types of ECG signals and a real-world ECG dataset to validate the effectiveness of the proposed methods. The experimental results demonstrate its superiority over representative state-of-the-art methods.

%The rest of the paper is organized as follows. In Section~\ref{sec:ProblemFormulation}, we briefly formulate the ECG classification problem. Then, we present the proposed method in Section~\ref{sec:ProposedApproach}, followed by the experiments and analysis in Section~\ref{sec:Experiments}. Finally, we conclude the paper and discuss future work.

\section{Problem formulation}
\label{sec:ProblemFormulation}
Given a set of ECG signals and their corresponding labels, the target of a classification method is to predict their labels correctly. As depicted in Section~\ref{sec:Intro}, it usually consists of two sequential modules: feature extraction and classifier training. Once the classifier is obtained, it can be used for unseen samples prediction, $i.e.$, testing phase. Mathematically, we denote the set of ECG wave signals as:
\begin{equation}
X = \left\{ {\left( {{x_i},{y_i}} \right)\left| {i \in \Lambda } \right.} \right\},
\label{eq:ECGWaveSignal}
\end{equation}
where ${x_i}$ is the $i^{th}$ sequence with $N$ samples : ${x_i} = {[{x_i}\left( 0 \right),{x_i}\left( 1 \right),...,{x_i}\left( {N - 1} \right)]^T} \in {R^N}$. ${y_i} \in \left\{ {0,..,C - 1} \right\}$ is the category of ${x_i}$, and $C$ is the number of total categories. $\Lambda $ is the index set of all samples. The feature extraction can be described as follows:
\begin{equation}
{f_i} = f\left( {{x_i},{\theta _f}} \right),
\label{eq:featureExtraction}
\end{equation}
where ${f_i} \in {R^M}$ is the corresponding feature representation of signal ${x_i}$. Usually, the feature vector ${f_i}$ is more compact than the original signal ${x_i}$, $i.e.$, $M \ll N$. $f\left( { \cdot ,{\theta _f}} \right)$ is a mapping function from the original signal space to the feature space, and ${\theta _f}$ is the parameters associated with the mapping $f\left(  \cdot  \right)$. It is usually determined according to domain knowledge of experts and cross-validation. Given the feature representation, a classifier $g\left( { \cdot ,{\theta _g}} \right)$ predicts its category as follows:
\begin{equation}
{c_i} = g\left( {f\left( {{x_i},{\theta _f}} \right),{\theta _g}} \right),
\label{eq:classifier}
\end{equation}
where ${\theta _g}$ is the parameters associated with the classifier $g\left(  \cdot  \right)$. ${c_i} \in \left\{ {0,..,C - 1} \right\}$ is the prediction. The frequently-used classifiers include SVM \cite{osowski2004support}, ANN \cite{barni2011privacy, mar2011optimization, wang2013ecg}, Random Forest, Deep CNN \cite{kiranyaz2016real}, $etc$. Given the training samples, the training of a classifier can be formulated as an optimization problem of its parameter ${\theta _g}$ as follows:
\begin{equation}
\theta _g^* = \mathop {\arg \min }\limits_{{\theta _g}} \sum\limits_{i \in {\Lambda _T}} {L\left( {g\left( {f\left( {{x_i},{\theta _f}} \right),{\theta _g}} \right),{y_i}} \right)},
\label{eq:thetaOpt}
\end{equation}
where ${\Lambda _T}$ is the index set of training samples. $L\left(  \cdot  \right)$ is a loss function which depicts the loss of assigning a prediction category ${c_i}$ for a sample ${x_i}$ with label ${y_i}$, $e.g.$, margin loss in SVM model and cross-entropy loss in models of ANN or Random Forest.

\begin{figure*}
\centering
\includegraphics[width=0.9\linewidth]{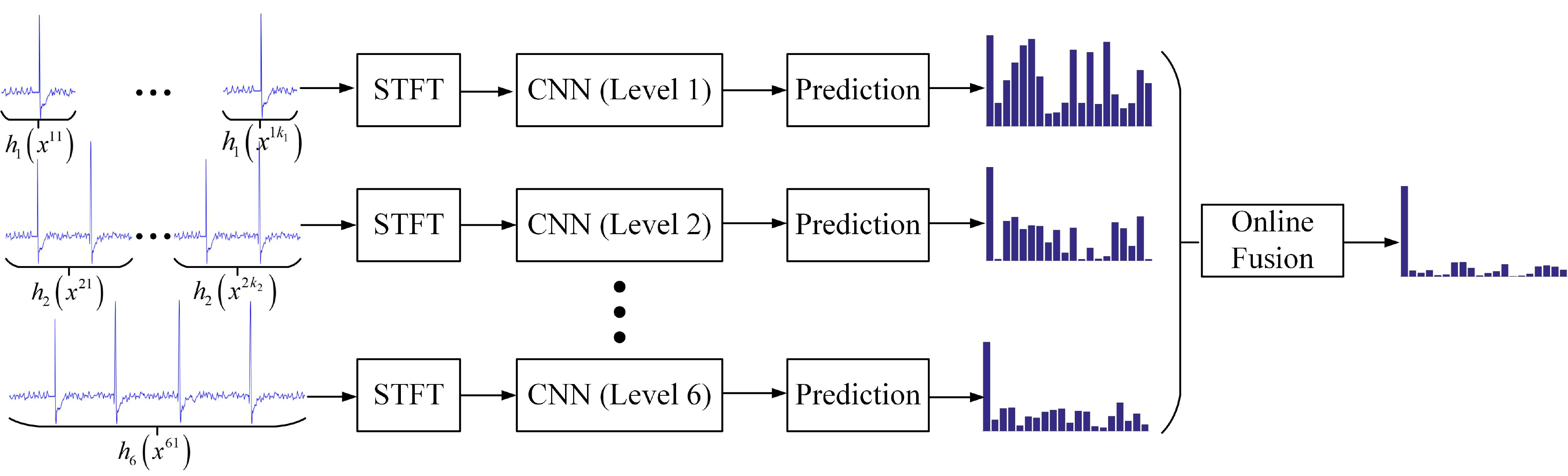}
\caption{The pipeline of the proposed method for ECG classification.}
\label{fig:flowchart}
\end{figure*}

\begin{figure*}
\centering
\includegraphics[width=0.8\linewidth]{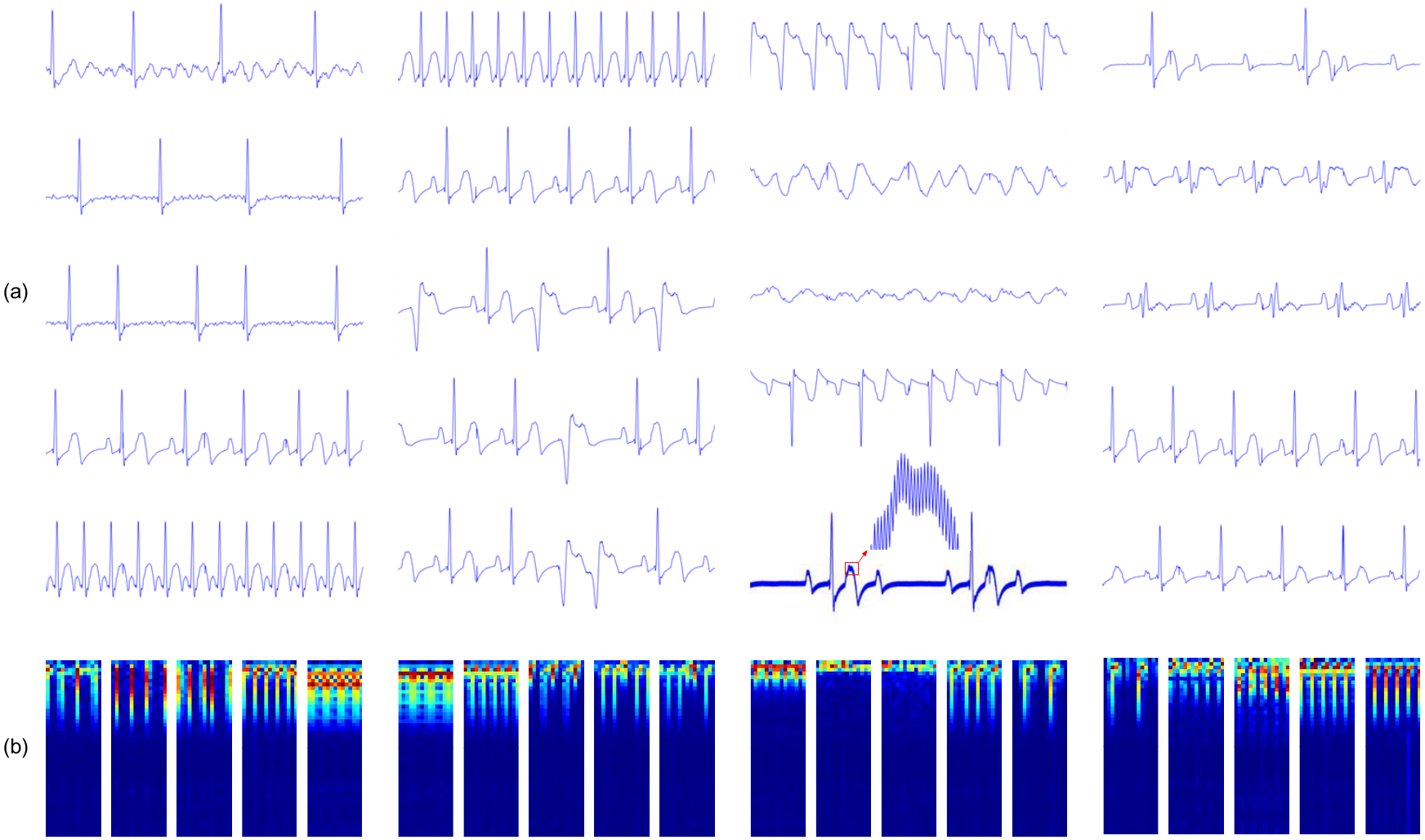}
\caption{(a) Exemplar original ECG wave signals from each category. (b) Spectrograms of (a) using Short-Time Fourier Transform.}
\label{fig:waveSpectrogram}
\end{figure*}

For deep neural networks models, feature extraction (learning) and classifier training are integrated together in the neural network architecture as an end-to-end model. The parameters are optimized for training samples by using the error back propagation algorithm. Mathematically, it can be formulated as:
\begin{equation}
\theta _h^* = \mathop {\arg \min }\limits_{{\theta _h}} \sum\limits_{i \in {\Lambda _T}} {L\left( {h\left( {{x_i},{\theta _h}} \right),{y_i}} \right)},
\label{eq:thetaOptDNN}
\end{equation}
where $h\left( { \cdot ,{\theta _h}} \right)$ is the deep neural networks model with parameters ${\theta _h}$. For a modern deep neural networks architecture, $e.g.$, Deep CNN, it usually consists of many sequential layers like convolutional layers, pooling layer, nonlinear activation layer, and fully connected layer, $etc$. Therefore, $h\left( { \cdot ,{\theta _h}} \right)$ is a nonlinear mapping function with strong representation capacity and maps the original high-dimension input data to a low-dimension feature space, where features are more discriminative and compact.

\section{The proposed approach}
\label{sec:ProposedApproach}

\subsection{Short-Time Fourier Transform}
\label{subsec:STFT}

Although wave signals in the original time domain can be used as input of DNN to learn features, a time-frequency representation calculated within a short-window may be a better choice \cite{ullah2010dna}. Inspired by the work in speech recognition areas \cite{deng2013recent}, where they show spectrogram features of speech are superior to Mel Frequency Cepstrum Coefficient (MFCC) with DNN, we first transform the original ECG wave signal into the time-frequency domain by using Short-Time Fourier Transform to obtain the ECG spectrogram representation. Mathematically, it can be described as follows:
\begin{equation}
{s_i}\left( {k,m} \right) = \sum\limits_{n = 0}^{N - 1} {{x_i}\left( n \right)w\left( {m - n} \right){e^{ - j\frac{{2\pi }}{N}kn}}},
\label{eq:STFT}
\end{equation}
where $w( \cdot )$ is the window function, $e.g.$, Hamming window. ${s_i}\left( {k,m} \right)$ is the two-dimension spectrogram of ${x_i}$. Figure~\ref{fig:waveSpectrogram} shows some examples of spectrograms.

\subsection{Architecture of the proposed CNN}
\label{subsec:Architecture}

\begin{table*}[htbp]
\small
  \centering
  \caption{The proposed network architectures for ECG classification.}
  %  \begin{tabular}{l|cccccc}
   \begin{tabular}{p{2.8cm}<{\centering}p{1.5cm}<{\centering}p{1.3cm}<{\centering}p{0.9cm}<{\centering}p{0.9cm}<{\centering}p{0.9cm}<{\centering}p{0.9cm}<{\centering}}
   \toprule
    Network & Type  & Input Size & Channels   & Filter & Pad   & stride \\
    \hline
    \multirow{8}[0]{*}{Proposed Network ($h1$)} & Conv1 & 1x32x4 & 32   & 3x3   & 1     & 1 \\
          & Pool1 & 32x32x4 & -     & 4x1     & 0     & (4,1) \\
          & Conv2 & 32x8x4 & 32     & 3x3   & 1     & 1 \\
          & Pool2 & 32x8x4 & -    & 4x2   & 0     & (4,2) \\
          & Fc3 & 32x2x2 & 64     & -   & -     & - \\
          & Fc4 & 64x1x1 & 20     & -   & -     & - \\
          & Params & \multicolumn{5}{c}{\textbf{18,976}}         \\
          & Complexity\footnotemark[1] & \multicolumn{5}{c}{3.5x10$^5$}         \\
    \midrule
    \multirow{8}[0]{*}{Network in \cite{kiranyaz2016real}} & Conv1 & 1x1x512 & 32   & 1x15   & (0,7)     & (1,6) \\
          & Conv2 & 32x1x86 & 16     & 1x15   & (0,7)     & (1,6) \\
          & Conv3 & 16x1x15 & 16     & 1x15   & (0,7)     & (1,6) \\
          & Pool3 & 16x1x3 & -     & 1x3     & 0     & 1 \\
          & Fc4 & 16x1x1 & 10    & -   & -     & - \\
          & Fc5 & 10x1x1 & 20     & -   & -     & - \\
          & Params & \multicolumn{5}{c}{\textbf{12,360}}         \\
          & Complexity & \multicolumn{5}{c}{1.7x10$^5$}         \\
          \bottomrule
    \end{tabular}
  \label{tab:Architecture}%
\end{table*}%

\footnotetext[1]{Evaluated with FLOPs, $i.e.$, the number of floating-point multiplication-adds.}

Since we use the two-dimension spectrogram as input, we design a deep CNN architecture that involves 2-D convolutions. Specifically, the proposed architecture consists of 3 convolutional layers and 2 fully-connected layers. There is a max-pooling layer and a ReLU layer after the first two convolutional layers and a max-pooling layer after the last convolutional layer, respectively. Details are shown in Table~\ref{tab:Architecture}. As can be seen, it is quite light-weight with 18,976 parameters and 3.5x10$^5$ FLOPs. We also present the network architecture in \cite{kiranyaz2016real} as a comparison. Filter sizes and strides are adapted to the data length used in this paper. It can be seen that the proposed network has a comparable amount of parameters and computational cost with the one in \cite{kiranyaz2016real}. As will be shown in Section~\ref{sec:Experiments}, the proposed method is computationally efficient and achieves a real-time performance even in an embedded device.

With the spectrogram ${s_i}$ as input, the CNN model predicts a probability vector ${p_i} = h\left( {{s_i},{\theta _h}} \right) \in {R^C}$ subjected to $\sum\limits_{c = 0}^{C - 1} {{p_{ic}}}  = 1$ and ${p_{ic}} \ge 0$. Then the model parameter ${\theta _h}$ can be learned by minimizing the cross-entropy loss as follows:
\begin{equation}
\begin{aligned}
{\theta _h^{*}} &= \mathop {\arg \min }\limits_{{\theta _h}} \sum\limits_{i \in {\Lambda _T}} {L\left( {h\left( {{s_i},{\theta _h}} \right),{y_i}} \right)} \\
 &= \mathop {\arg \min }\limits_{{\theta _h}}  - \sum\limits_{i \in {\Lambda _T}} {\sum\limits_{c = 0}^{C - 1} {{q_{ic}}\log \left( {{p_{ic}}} \right)} },
 \end{aligned}
\label{eq:thetaOptCE}
\end{equation}
where ${q_i} \in {R^C}$ is the one-hot vector of label ${y_i}$, $i.e.$, ${q_{ic}} = \delta \left( {{y_i},c} \right) \in \left\{ {0,1} \right\}$.

Usually, single beat is detected and classified \cite{guler2005ecg}. However, since long signals contain more beats given the sampling rate, prediction on it will be more accurate. In this paper, the length of each sample in the synthetic ECG dataset is 16384 at a sampling rate of 512Hz, which lasts 32s. We split each sample into sub-samples which have the same length of 512. Therefore, each sample contains about $\sim1$ beats. It is noteworthy that we do not explicitly extract the beat from the raw signal but use it as the input directly after the Short-Time Fourier Transform. Then, we train our CNN model on this dataset. Besides, to compare the performance of models for longer samples, we also split each sample into sub-samples of different lengths, $e.g.$, 2s, 4s, 8s, 16s. We use them to train our CNN models accordingly and denote all these scale-specific models as ${h_1} \sim {h_6}$, respectively. The width of the spectrogram is determined by the length of the wave signal given the window function, which varies for each model. Nevertheless, we use the same architecture for all models and change the pooling strides along columns accordingly while keeping the fully-connected layers fixed.

\subsection{Optimization}
\label{subsec:Optimization}
The optimization of Eq.~\eqref{eq:thetaOptCE} is not trivial since the objective function $L\left(  \cdot  \right)$ is non-linear and non-convex. Instead of using deterministic optimization methods \cite{jones1993lipschitzian,sergeyev2018efficiency}, we adopt the mini-batch Stochastic Gradient Descent (SGD) algorithm \cite{bottou1998online, bottou2010large} in this paper. Mathematically, it can be formulated as:
\begin{equation}
\theta _h^{t + 1} = \theta _h^t - {\gamma _t}\frac{1}{B}\sum\limits_{i = 1}^B {{\nabla _{{\theta _h}}}L\left( {h\left( {{s_i},{\theta _h}} \right),{y_i}} \right)},
\label{eq:sgd}
\end{equation}
where $B$ is the number of training samples in each mini-batch, $\gamma _t$ is the learning rate at step $t$. It can be proved that as long as the learning rate $\gamma _t$ are small enough, the algorithm converges towards a local minimum of the empirical risk \cite{bottou1998online}. Specifically, to keep the optimization direction and prevent oscillations, we leverage the SGD algorithm with a momentum term \cite{rumelhart1988learning,sutskever2013importance}, $i.e.$,
\begin{equation}
\Delta \theta _h^{t + 1} = \alpha \Delta \theta _h^t - {\gamma _t}\frac{1}{B}\sum\limits_{i = 1}^B {{\nabla _{{\theta _h}}}L\left( {h\left( {{s_i},{\theta _h}} \right),{y_i}} \right)},
\label{eq:sgd_momentum}
\end{equation}
and
\begin{equation}
\theta _h^{t + 1} = \theta _h^t + \Delta \theta _h^{t + 1},
\label{eq:sgd_momentum_update}
\end{equation}
where $\Delta \theta _h$ is the momentum term and $\alpha$ is the momentum parameter. The setting of $\alpha$, $\gamma$, and $B$ will be presented in Section~\ref{subsubsec:DatasetParams}.

\subsection{Progressive online decision fusion}
\label{subsec:OnlineFusion}
For online testing, as the length of signal is growing, we can test it sequentially by using the scale-specific models. As illustrated in Figure~\ref{fig:flowchart}, lower level models make decisions based on local patterns within short signals, while higher level models make decisions based on global patterns within long signals. These models can be seen as different experts focusing on different scales, whose decisions are complementary and could be fused as a more accurate and stable one \cite{zhang2015multi,lu2016multilevel}. To this end, we propose a progressive online decision fusion method. Mathematically, it can be described as follows:
\begin{equation}
\widetilde{h}\left( x \right) = \sum\limits_{s = 1}^{s = {s_l}} {{w_s}\left( {\frac{1}{{{k_s}}}\sum\limits_{k = 1}^{k = {k_s}} {{h_s}\left( {{x^{sk}}} \right)} } \right)},
\label{eq:fusion}
\end{equation}
where $\widetilde{h}\left(  \cdot  \right)$ represents the fusion result, ${s_l} \in \left\{ {1,2,3,4,5,6} \right\}$ is the maximum level for a signal $x$ of specific length. ${x^{sk}}$ is the ${k^{th}}$ segment of $x$ for the ${s^{th}}$ level model ${h_s}$, and ${k_s}$ is the number of segments at $s^{th}$ level, $i.e.$, $k_s = 2^{s_l - s}$. For example, when the length of $x$ is 2048, ${s_l}$ will be 3, and ${k_1} \sim {k_3}$ will be 4, 2, 1, respectively. ${w_s}$ is the fusion weight of ${h_s}$ subjected to $\sum\limits_{s = 1}^{s = {s_l}} {{w_s}}  = 1$. It can be seen from Eq.~\eqref{eq:fusion}, decision at each segment at the same level is treated equally. It is reasonable since there is no prior knowledge favouring specific segment and the decision is made by the same model.

Assuming that the distribution of $h_s\left(  \cdot  \right)$ is independent from each other for any $s$, with a mean $\mu _s$ and variance $\sigma _s$, the expectation of Eq.~\eqref{eq:fusion} can be derived as:
\begin{equation}
\begin{aligned}
E\left [ \widetilde{h}\left( x \right) \right ] & = \sum\limits_{s = 1}^{s = {s_l}} {{w_s}\left( {\frac{1}{{{k_s}}}\sum\limits_{k = 1}^{k = {k_s}} {E\left [ {h_s}\left( {{x^{sk}}} \right) \right ] } } \right)} \\
& = \sum\limits_{s = 1}^{s = {s_l}} {{w_s}\left( {\frac{1}{{{k_s}}}\sum\limits_{k = 1}^{k = {k_s}} {\mu _s} } \right)} \\
& = \sum\limits_{s = 1}^{s = {s_l}} {{w_s \mu _s}}.
\end{aligned}
\label{eq:expectation}
\end{equation}
For a given training sample $x_i$, $h_s\left(  \cdot  \right)$ shares the same training targets $y_i$. Therefore, $\mu _s$ should be the same for any $s$, denoted as $\mu$. Accordingly, we have the unbiased estimate $E\left [ \widetilde{h}\left( x \right) \right ] = \mu$. Similarly, we can derive the variance of $\widetilde{h}\left( x \right)$ as: $Var \left [ \widetilde{h}\left( x \right) \right ] = \sum\limits_{s = 1}^{s = {s_l}} {{\frac{w_s^2}{k_s} \sigma _s}}$. Usually, the variance $\sigma _s$ is decreased with the growth of sample length (scale $s$) since more ``evidence'' is accumulated. For instance, if we assume $\sigma _s = \frac{1}{k_s}\sigma$, then we will have $Var \left [ \widetilde{h}\left( x \right) \right ] =  \frac{2^{2s_l + 2} -2^2}{2^{2s_l}s_l^2\left (2^2-1\right )} \sigma \le \sigma$ for uniform weights $w_s=\frac{1}{s_l}$, and $Var \left [ \widetilde{h}\left( x \right) \right ] =  \frac{2^{4s_l + 2} -2^2}{2^{2s_l} \left (2^{s_l}-1\right )^2 \left ( 2^4-1 \right ) } \sigma \le \sigma$ for non-uniform weights defined in Eq.~\eqref{eq:weight}. Taking $s_l=6$ as an example, we will have $Var \left [ \widetilde{h}\left( x \right) \right ] \approx 0.037\sigma$ and $Var \left [ \widetilde{h}\left( x \right) \right ] \approx 0.2752\sigma$ for uniform and non-uniform weights, respectively. As can be seen, using a fusion decision will reduce the variance and get a more stable result.

\section{Experiments}
\label{sec:Experiments}

\subsection{Evaluation metrics}
\label{subsec:evaluationMetrics}
First, we present the definition of the evaluation metrics used in this paper. Denoting the confusion matrix as $CM = \left[ {{c_{ij}}} \right]$, where $c_{ij}$ is number of samples belonging to the $i^{th}$ category but being predicted as the $j^{th}$ one, the Accuracy, Sensitivity, Specificity and F1 score can be calculated as follows.

\begin{equation}
Accuracy = {{\sum\limits_{i = 1}^{i = C} {{c_{ii}}} } \mathord{\left/
 {\vphantom {{\sum\limits_{i = 1}^{i = C} {{c_{ii}}} } {\sum\limits_{i = 1}^{i = C} {\sum\limits_{j = 1}^{j = C} {{c_{ij}}} } }}} \right.
 \kern-\nulldelimiterspace} {\sum\limits_{i = 1}^{i = C} {\sum\limits_{j = 1}^{j = C} {{c_{ij}}} } }}.
\label{eq:accuracyScore}
\end{equation}

\begin{equation}
Sensitivit{y_i} = {{{c_{ii}}} \mathord{\left/
 {\vphantom {{{c_{ii}}} {\sum\limits_{j = 1}^{j = C} {{c_{ij}}} }}} \right.
 \kern-\nulldelimiterspace} {\sum\limits_{j = 1}^{j = C} {{c_{ij}}} }},
\label{eq:sensitivityScore}
\end{equation}
where class $i$ represents the symptomatic classes, $e.g.$, RAF, FAF, $etc$.

\begin{equation}
Specificit{y_k} = {{{c_{kk}}} \mathord{\left/
 {\vphantom {{{c_{kk}}} {\sum\limits_{j = 1}^{j = C} {{c_{kj}}} }}} \right.
 \kern-\nulldelimiterspace} {\sum\limits_{j = 1}^{j = C} {{c_{kj}}} }},
\label{eq:specifictyScore}
\end{equation}
where class $k$ represents the normal class.

\begin{equation}
F{1_i} = {{2{c_{ii}}} \mathord{\left/
 {\vphantom {{2{c_{ii}}} {\left( {\sum\limits_{j = 1}^{j = C} {{c_{ij}}}  + \sum\limits_{i = 1}^{i = C} {{c_{ij}}} } \right)}}} \right.
 \kern-\nulldelimiterspace} {\left( {\sum\limits_{j = 1}^{j = C} {{c_{ij}}}  + \sum\limits_{i = 1}^{i = C} {{c_{ij}}} } \right)}}.
\label{eq:F1Score}
\end{equation}

\begin{equation}
F1 = \frac{1}{C}\sum\limits_{i = 1}^{i = C} {F{1_i}}.
\label{eq:meanF1Score}
\end{equation}

% Table generated by Excel2LaTeX from sheet 'Sheet3'
\begin{table*}[htbp]
  \scriptsize
  \centering
  \caption{Sensitivity and Specificity scores of different methods on the training set. Mean scores and the Standard deviations (mean$\pm$std) are reported.}
   \setlength{\tabcolsep}{1.5mm}
    \begin{tabular}{cccccccc}
    %\begin{tabular}{p{2.8cm}<{\centering}p{0.7cm}<{\centering}p{0.7cm}<{\centering}p{0.7cm}<{\centering}p{0.7cm}<{\centering}p{0.7cm}<{\centering}p{0.7cm}<{\centering}p{0.7cm}<{\centering}p{0.7cm}<{\centering}p{0.7cm}<{\centering}p{1.1cm}<{\centering}}
    \toprule
          & \multicolumn{5}{c}{Sensitivity}                                               \\
    \midrule
       Methods   & RAF   & FAF   & AF    & SA    & AT   & ST    & PAC  \\
          \midrule
    SVM+FFT & 0.95$\pm$0.02  & 0.73$\pm$0.01  & 0.77$\pm$0.08  & \textbf{0.77$\pm$0.05}  & 0.85$\pm$0.02  & 0.78$\pm$0.10  & 0.76$\pm$0.08  \\
    1D CNN \cite{kiranyaz2016real} & 0.96$\pm$0.01  & 0.92$\pm$0.04  & 0.84$\pm$0.06  & 0.39$\pm$0.03  & 0.90$\pm$0.01  & 0.96$\pm$0.02  & 0.20$\pm$0.09 \\
    Proposed & 0.97$\pm$0.01  & 0.99$\pm$0.01  & 0.94$\pm$0.01  & 0.71$\pm$0.02  & 0.94$\pm$0.01  & 0.99$\pm$0.01  & 0.59$\pm$0.20 \\
    SVM+CNN Feature & \textbf{0.98$\pm$0.01}  & \textbf{0.99$\pm$0.01}  & \textbf{0.97$\pm$0.01} & 0.75$\pm$0.01  & \textbf{0.97$\pm$0.01}   & \textbf{0.99$\pm$0.01}  & \textbf{0.82$\pm$0.04}\\

    \toprule
          & \multicolumn{5}{c}{Sensitivity}                                               \\
    \midrule
        Methods     & VB    & VTr   & PVCCI  & VTa   & RVF   & FVF   & AVB-I  \\
           \midrule
     SVM+FFT    & 0.81$\pm$0.04  & \textbf{0.71$\pm$0.10}  & 0.82$\pm$0.03  & 0.84$\pm$0.02  & 0.86$\pm$0.03  & 0.78$\pm$0.06  & 0.82$\pm$0.02\\
     1D CNN \cite{kiranyaz2016real}     & 0.92$\pm$0.03  & 0.19$\pm$0.11  & 0.87$\pm$0.02  & 0.97$\pm$0.01  & 0.97$\pm$0.02  & 0.96$\pm$0.02  & 0.70$\pm$0.17 \\
     Proposed    & 0.93$\pm$0.01  & 0.18$\pm$0.19  & 0.89$\pm$0.03  & 0.99$\pm$0.01  & 0.99$\pm$0.01  & 0.98$\pm$0.01  & 0.77$\pm$0.15 \\
     SVM+CNN Feature   & \textbf{0.96$\pm$0.01}  & 0.34$\pm$0.12  & \textbf{0.95$\pm$0.01} & \textbf{1.00$\pm$0.00}     & \textbf{1.00$\pm$0.00}     & \textbf{0.99$\pm$0.01}  & \textbf{0.95$\pm$0.01} \\
     \midrule
          & \multicolumn{5}{c}{Sensitivity}                                       & Specificity \\
     \midrule
       Methods   & AVB-II & AVB-III & RBBB  & LBBB  & PVC   & N \\
     \midrule
    SVM+FFT  & 0.84$\pm$0.05  & 0.76$\pm$0.07  & 0.72$\pm$0.09  & 0.81$\pm$0.05  & \textbf{0.76$\pm$0.10}  & 0.86$\pm$0.02 \\
    1D CNN \cite{kiranyaz2016real}   & 0.94$\pm$0.03  & 0.68$\pm$0.05  & 0.91$\pm$0.04  & 0.98$\pm$0.01  & 0.68$\pm$0.05  & 0.95$\pm$0.01 \\
    Proposed   & 0.92$\pm$0.03  & 0.86$\pm$0.04  & 0.95$\pm$0.01  & 0.98$\pm$0.01  & 0.72$\pm$0.05  & 0.96$\pm$0.01 \\
    SVM+CNN Feature   & \textbf{0.98$\pm$0.01}  & \textbf{0.93$\pm$0.03}  & \textbf{0.97$\pm$0.01}  & \textbf{0.98$\pm$0.01}  & 0.75$\pm$0.04  & \textbf{0.98$\pm$0.01} \\
    \bottomrule
    \end{tabular}%
  \label{tab:SensitivitySpecificityTrain}%
\end{table*}%

% Table generated by Excel2LaTeX from sheet 'Sheet3'
\begin{table*}[htbp]
  \scriptsize
  \centering
  \caption{Sensitivity and Specificity scores of different methods on the test set. Mean scores and the Standard deviations (mean$\pm$std) are reported.}
    \setlength{\tabcolsep}{1.5mm}
    \begin{tabular}{cccccccc}
    %\begin{tabular}{p{2.8cm}<{\centering}p{0.7cm}<{\centering}p{0.7cm}<{\centering}p{0.7cm}<{\centering}p{0.7cm}<{\centering}p{0.7cm}<{\centering}p{0.7cm}<{\centering}p{0.7cm}<{\centering}p{0.7cm}<{\centering}p{0.7cm}<{\centering}p{1.1cm}<{\centering}}
    \toprule
          & \multicolumn{5}{c}{Sensitivity}                                               \\
    \midrule
      Methods    & RAF   & FAF   & AF    & SA    & AT    & ST    & PAC  \\
          \midrule
    SVM+FFT & 0.88$\pm$0.11  & 0.44$\pm$0.14  & 0.50$\pm$0.21   & 0.69$\pm$0.13  & 0.53$\pm$0.27  & 0.56$\pm$0.21  & 0.57$\pm$0.13  \\
    1D CNN \cite{kiranyaz2016real} & 0.96$\pm$0.01  & 0.93$\pm$0.04  & 0.83$\pm$0.07  & 0.38$\pm$0.04  & 0.90$\pm$0.02   & 0.95$\pm$0.03  & 0.20$\pm$0.05 \\
    Proposed & 0.98$\pm$0.01  & 0.99$\pm$0.01  & 0.95$\pm$0.03  & 0.72$\pm$0.02  & 0.94$\pm$0.02  & 0.96$\pm$0.03  & 0.62$\pm$0.22   \\
    SVM+CNN Feature & 0.97$\pm$0.01  & 0.98$\pm$0.01  & 0.96$\pm$0.01  & 0.72$\pm$0.02  & 0.95$\pm$0.01  & 0.96$\pm$0.04  & 0.85$\pm$0.03   \\
    Proposed(Fusion) & \textbf{0.99$\pm$0.01}  & \textbf{1.00$\pm$0.00}     & \textbf{0.99$\pm$0.01}  & \textbf{1.00$\pm$0.00}     & \textbf{1.00$\pm$0.00}     & \textbf{1.00$\pm$0.00}     & \textbf{1.00$\pm$0.00}   \\
   \toprule
          & \multicolumn{5}{c}{Sensitivity}                                               \\
    \midrule
        Methods   & VB    & VTr   & PVCCI   & VTa   & RVF   & FVF   & AVB-I \\
    SVM+FFT & 0.42$\pm$0.29  & 0.38$\pm$0.14  & 0.54$\pm$0.24 & 0.73$\pm$0.09  & 0.60$\pm$0.18   & 0.45$\pm$0.32  & 0.50$\pm$0.30   \\
    1D CNN \cite{kiranyaz2016real}  & 0.93$\pm$0.02  & 0.15$\pm$0.15  & 0.85$\pm$0.03  & 0.96$\pm$0.01  & 0.96$\pm$0.02  & 0.94$\pm$0.04  & 0.50$\pm$0.35  \\
    Proposed & 0.94$\pm$0.01  & 0.32$\pm$0.23  & 0.91$\pm$0.02  & 0.99$\pm$0.01  & 0.98$\pm$0.02  & 0.99$\pm$0.01  & 0.56$\pm$0.30  \\
    SVM+CNN Feature & 0.96$\pm$0.01  & 0.02$\pm$0.02 & 0.95$\pm$0.03  & 0.98$\pm$0.01  & 0.97$\pm$0.02  & 0.98$\pm$0.02  & 0.48$\pm$0.15  \\
    Proposed(Fusion) & \textbf{1.00$\pm$0.00}     & \textbf{0.99$\pm$0.01}  & \textbf{0.99$\pm$0.01} & \textbf{1.00$\pm$0.00}     & \textbf{1.00$\pm$0.00}     & \textbf{0.99$\pm$0.01}  & \textbf{0.86$\pm$0.20}  \\
    \midrule
          & \multicolumn{5}{c}{Sensitivity}                                       & Specificity \\
    \midrule
      Methods    & AVB-II & AVB-III & RBBB  & LBBB  & PVC   & N \\
    \midrule
    SVM+FFT   & 0.66$\pm$0.14  & 0.56$\pm$0.21  & 0.46$\pm$0.07  & 0.48$\pm$0.33  & 0.40$\pm$0.24   & 0.60$\pm$0.18 \\
    1D CNN \cite{kiranyaz2016real}   & 0.93$\pm$0.08  & 0.64$\pm$0.07  & 0.89$\pm$0.10  & 0.98$\pm$0.01  & 0.69$\pm$0.05  & 0.95$\pm$0.02 \\
    Proposed   & 0.93$\pm$0.06  & 0.87$\pm$0.05  & 0.95$\pm$0.02  & 0.98$\pm$0.01  & 0.57$\pm$0.25  & 0.96$\pm$0.02 \\
    SVM+CNN Feature   & 0.95$\pm$0.05  & 0.89$\pm$0.05  & 0.95$\pm$0.02  & 0.98$\pm$0.01  & 0.57$\pm$0.22  & 0.97$\pm$0.02 \\
    Proposed(Fusion)   & \textbf{1.00$\pm$0.00}     & \textbf{0.95$\pm$0.05}  & \textbf{0.98$\pm$0.02}  & \textbf{1.00$\pm$0.00}     & \textbf{0.99$\pm$0.01}  & \textbf{0.99$\pm$0.01} \\
    \bottomrule
    \end{tabular}%
  \label{tab:SensitivitySpecificityVal}%
\end{table*}%

\subsection{Experiments on a synthetic ECG dataset}
\label{subsec:syntheticExperiments}

\subsubsection{Dataset and parameter settings}
\label{subsubsec:DatasetParams}

To verify the effectiveness of the proposed method, we construct a synthetic dataset by using an ECG simulator. The simulator can generate different types of ECG signals with different parameter settings. Generally, the ECG signals consist of four types of features, namely, trend, cycle, irregularities, and burst \cite{ullah2019modeling}. To cover these features, we choose 20 categories of ECG signals in this paper, which include Normal (N), Rough Atrial Fibrillation (RAF), Fine Atrial Fibrillation (FAF), Atrial Flutter (AF), Sinus Arrhythmia (SA), Atrial Tachycardia (AT), Supraventricular Tachycardia (ST), Premature Atrial Contraction (PAC), Ventricular Bigeminy (VB), Ventricular Trigeminy (VTr), Premature Ventricular Contraction Coupling Interval (PVCCI), Ventricular Tachycardia (VTa), Rough Ventricular Fibrillation (RVF), Fine Ventricular Fibrillation (FVF), Atrio-Ventricular Block I (AVB-I), Atrio-Ventricular Block (AVB-II), Atrio-Ventricular Block (AVB-III), Right Bundle Branch Block (RBBB), Left Bundle Branch Block (LBBB), and Premature Ventricular Contractions (PVC). There are a total of 2426 samples, about 120 samples per category. Each sample has a maximum length of 16384 points sampled at 512Hz. We split the sequence at a random position and merge them by changing their orders. In this way, we augment the dataset and simulate the time delay such that the model can capture the regularity and irregularity in the time series \cite{sharif2013fuzzy}. We use the 3-fold cross-validation to evaluate the proposed method.

Parameters are set as follows. We use the Hamming window of length 256 in Short-Time Fourier Transform and the overlap size is 128. The CNN model is trained in a total of 20,000 iterations with a batch size of 128. The learning rate decreases by half every 5,000 iterations from 0.01 to 6.25x$10^{-4}$. The momentum and the decay parameter are set to 0.9 and 5x$10^{-6}$, respectively. We implement the proposed method in CAFFE \cite{jia2014caffe} on a workstation with NVIDIA GTX Titan X GPUs if not specified.

\subsubsection{Comparisons with previous methods}
\label{subsubsec:Comparisons}

We compare the performance of the proposed method with previous methods including SVM based on Fourier transform, the pilot Deep CNN method in \cite{kiranyaz2016real} which uses 1-D convolutions, and SVM based on the learned features of the proposed method. We report the sensitivity and specificity scores of different methods on both the training set and test set. We also report the average classification accuracy. The standard deviations of each index on the 3-fold cross-validation are also reported. Results are summarized in Table~\ref{tab:SensitivitySpecificityTrain}, Table~\ref{tab:SensitivitySpecificityVal} and Table~\ref{tab:Accuracy}.

% Table generated by Excel2LaTeX from sheet 'acc'
\begin{table}[htbp]
 \renewcommand\arraystretch{0.8}
  \centering
  \caption{Average classification accuracy of different methods on the training set and test test. Standard deviations (Std.) are listed in the brackets.}
    \begin{tabular}{ccc}
    \toprule
    Methods & Training Set & Test Set \\
    \midrule
    SVM+FFT & 0.81($\pm$0.04)  & 0.56($\pm$0.12) \\
    1D CNN \cite{kiranyaz2016real} & 0.83($\pm$0.01)  & 0.81($\pm$0.04) \\
    Proposed & 0.88($\pm$0.03)  & 0.87($\pm$0.03) \\
    SVM+CNN Feature & 0.93($\pm$0.01)  & 0.87($\pm$0.02) \\
      Proposed (Fusion)    & 0.99($\pm$0.01)    & \textbf{0.99($\pm$0.01)} \\
    \bottomrule
    \end{tabular}%
  \label{tab:Accuracy}%
\end{table}%

It can be seen that the method in \cite{kiranyaz2016real} outperforms the traditional method using the FFT coefficients and SVM classifier. However, it is inferior to the proposed one which uses the 2-dimensional spectrogram as input, which benefits from the learned features of time-frequency characteristics. Besides, we use the learned features from the proposed method to train an SVM classifier. The results are denoted as ``SVM+CNN Feature''. Compared with the SVM with FFT features, the performance of this classifier is significantly boosted. It demonstrates that the proposed method learns a more discriminative feature representation of the ECG signal. Interestingly, it is marginally better than the proposed CNN model which employs a linear classifier. It is reasonable since a more sophisticated nonlinear radial basis kernel is used in the SVM classifier. However, it shows a tendency toward overfitting $i.e.$, a larger gain on the training set.

Moreover, from Table~\ref{tab:SensitivitySpecificityTrain} and Table~\ref{tab:SensitivitySpecificityVal}, we can find that categories of SA, PAC, VTr, AVB-I, and PVC are hard to be distinguished. We'll shed light on the phenomenon by inspecting the learned features through the visualization technique and analyzing the confusion matrix between categories as follows.

\subsubsection{Analysis on learned features and confusion matrix between categories}
\label{subsubsec:AnalysisFeatures}
\begin{figure*}
\centering
\includegraphics[width=0.9\linewidth]{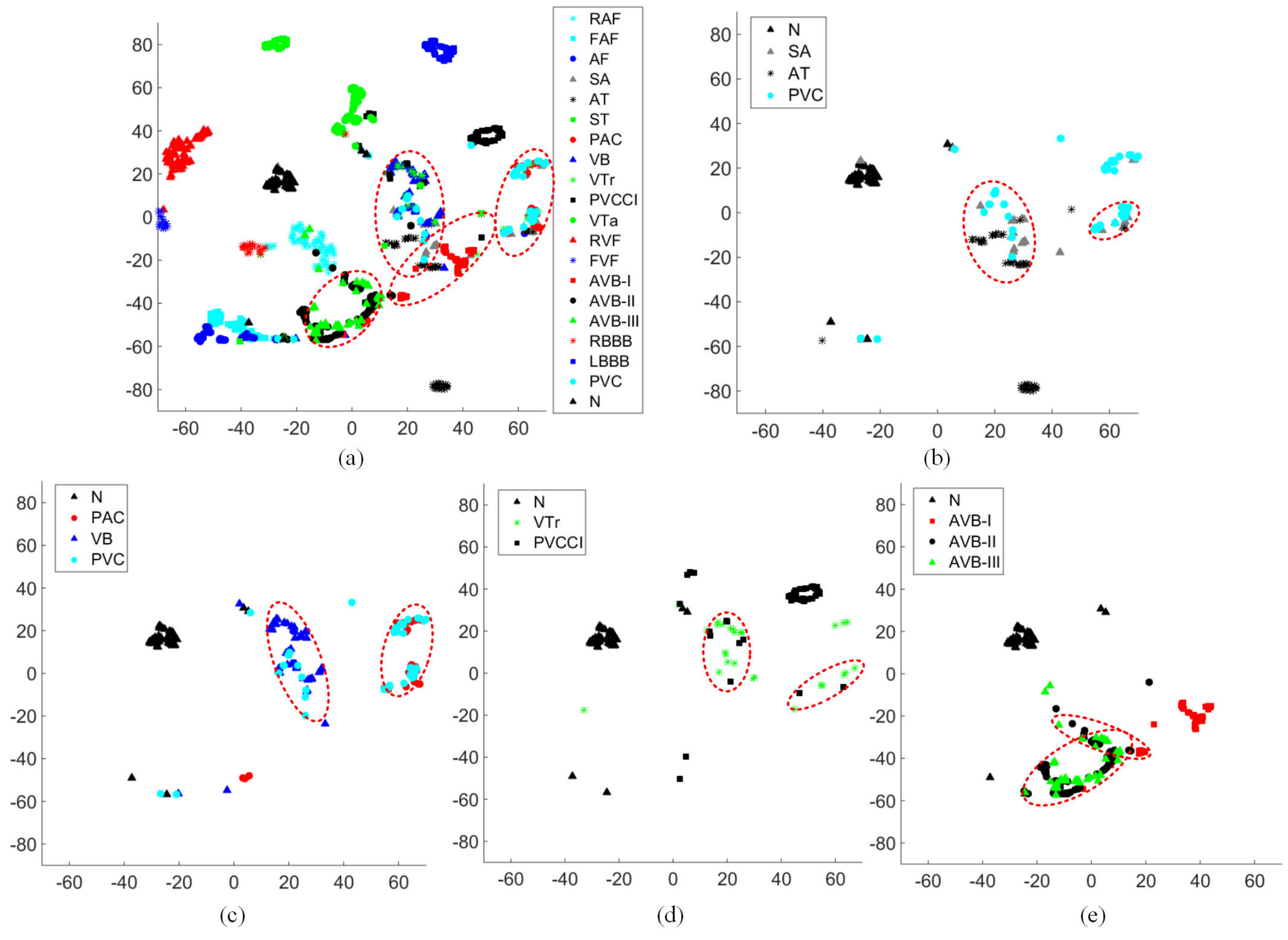}
\caption{Visualization of the learned features from the proposed network by using t-SNE \cite{maaten2008visualizing}. (a): Feature clusters of all categories on the test set. (b)-(e): Feature clusters of most confusion categories (denoted by the red circles). Best viewed on screen.}
\label{fig:featureProj}
\end{figure*}

First, we calculate the learned features from the penultimate layer for all test data. Then, we employ the t-Distributed Stochastic Neighbor Embedding (t-SNE) method proposed in \cite{maaten2008visualizing, van2014accelerating} to visually inspect them. The visualization results are shown in Figure~\ref{fig:featureProj}(a). As can be seen, some categories such as Normal(N), RVF, FVF, ST, RBBB, LBBB, PVCCI, VTa and RAF, are separated from other categories. However, some categories such as SA, PAC, PVC, VTr, AVB-I, and AVB-III, are overlapped with other categories as indicated by the red circles. We further plot them separately in Figure~\ref{fig:featureProj}(b)-(e). For example, SA tends to be overlapped with AT and PVC, and PVC tends to be overlapped with PAC and VB. Nevertheless, they are separated from the Normal category, coinciding with the high specificity scores in Table~\ref{tab:SensitivitySpecificityTrain} and Table~\ref{tab:SensitivitySpecificityVal}.

\begin{figure*}
\centering
\subfloat[]{\label{Fig:ConfusionMatrixLevel1}%%
\includegraphics[width=0.45\linewidth]{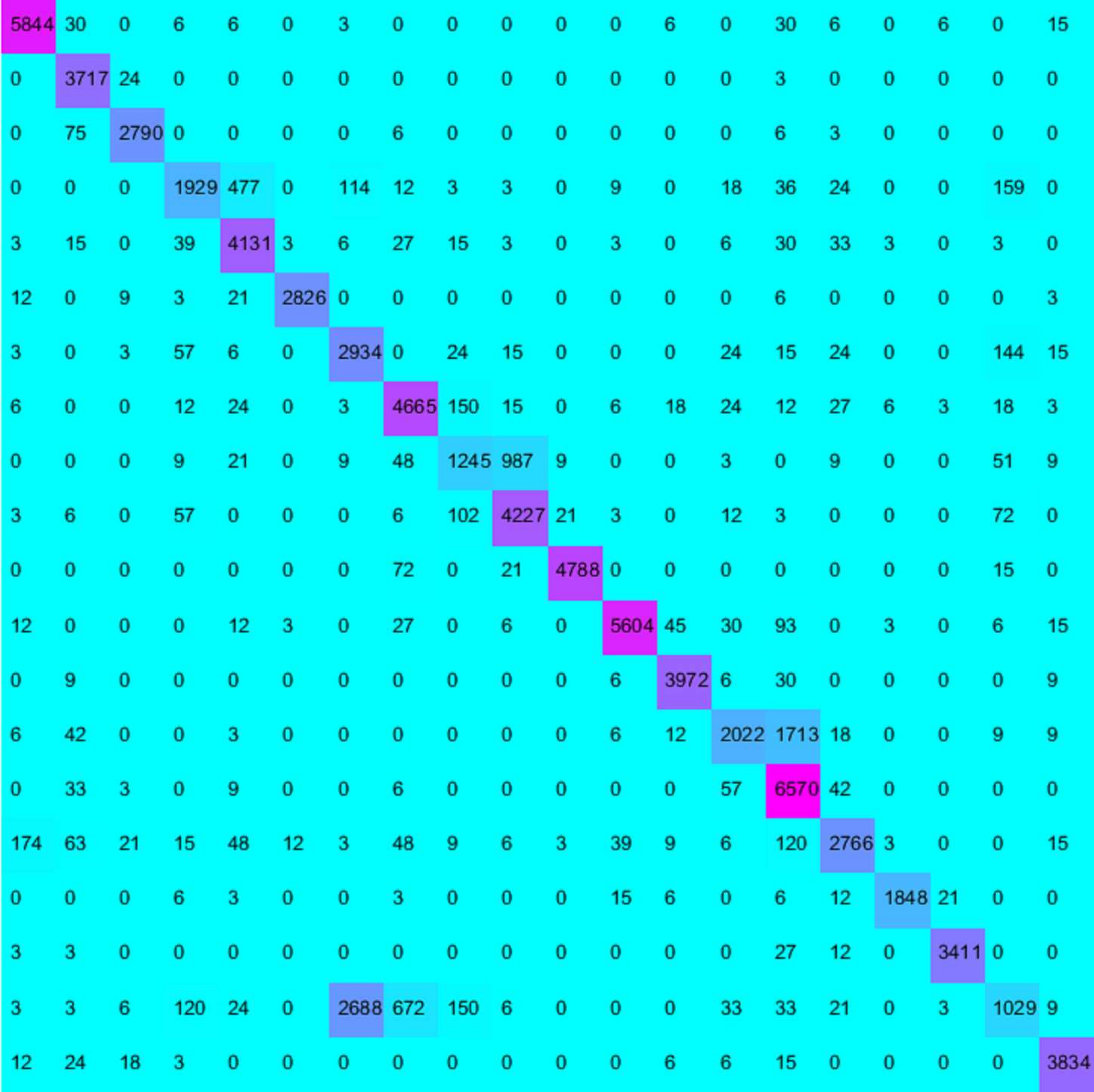}}
\hspace{0.025\linewidth}
\subfloat[]{\label{Fig:ConfusionMatrixFusion}%%
\includegraphics[width=0.45\linewidth]{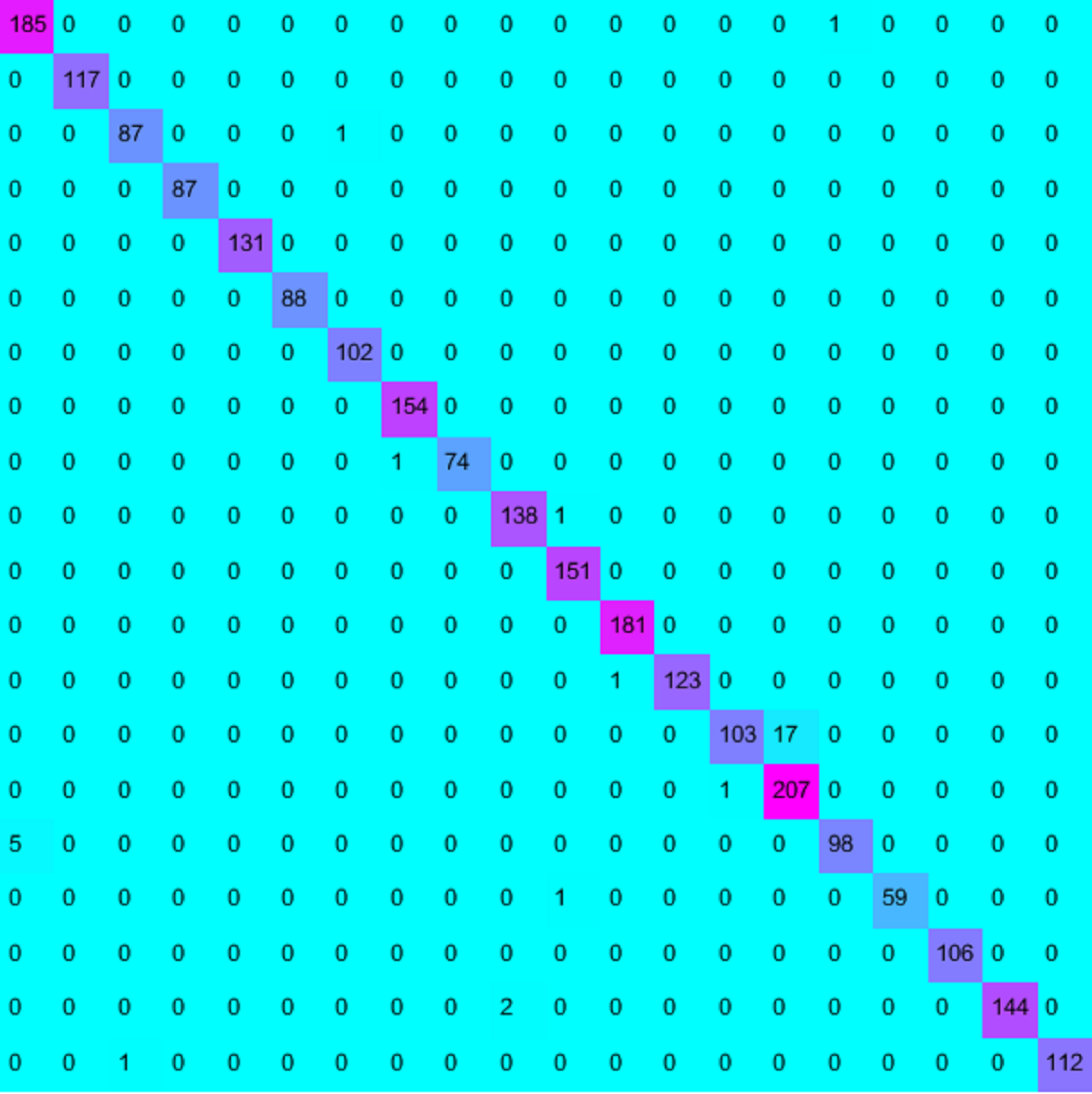}}

\caption{\protect\subref{Fig:ConfusionMatrixLevel1} Confusion matrix of the proposed method at level 1 on the test set. \protect\subref{Fig:ConfusionMatrixFusion} Confusion matrix of the proposed online decision fusion method on the test set. From top to down (top to down), the ground truth (predicted) categories are RAF, FAF, AF, SA, AT, ST, PAC, VB, VTr, PVCCI, VTa, RVF, FVF, AVB-I, AVB-II, AVB-III, RBBB, LBBB, PVC and N, respectively.}
\label{Fig:ConfusionMatrix}
\end{figure*}

\begin{figure*}
\centering
\subfloat[]{\label{Fig:accurayMultiScaleWholeAugFusionMu}%%
\includegraphics[width=0.39\linewidth]{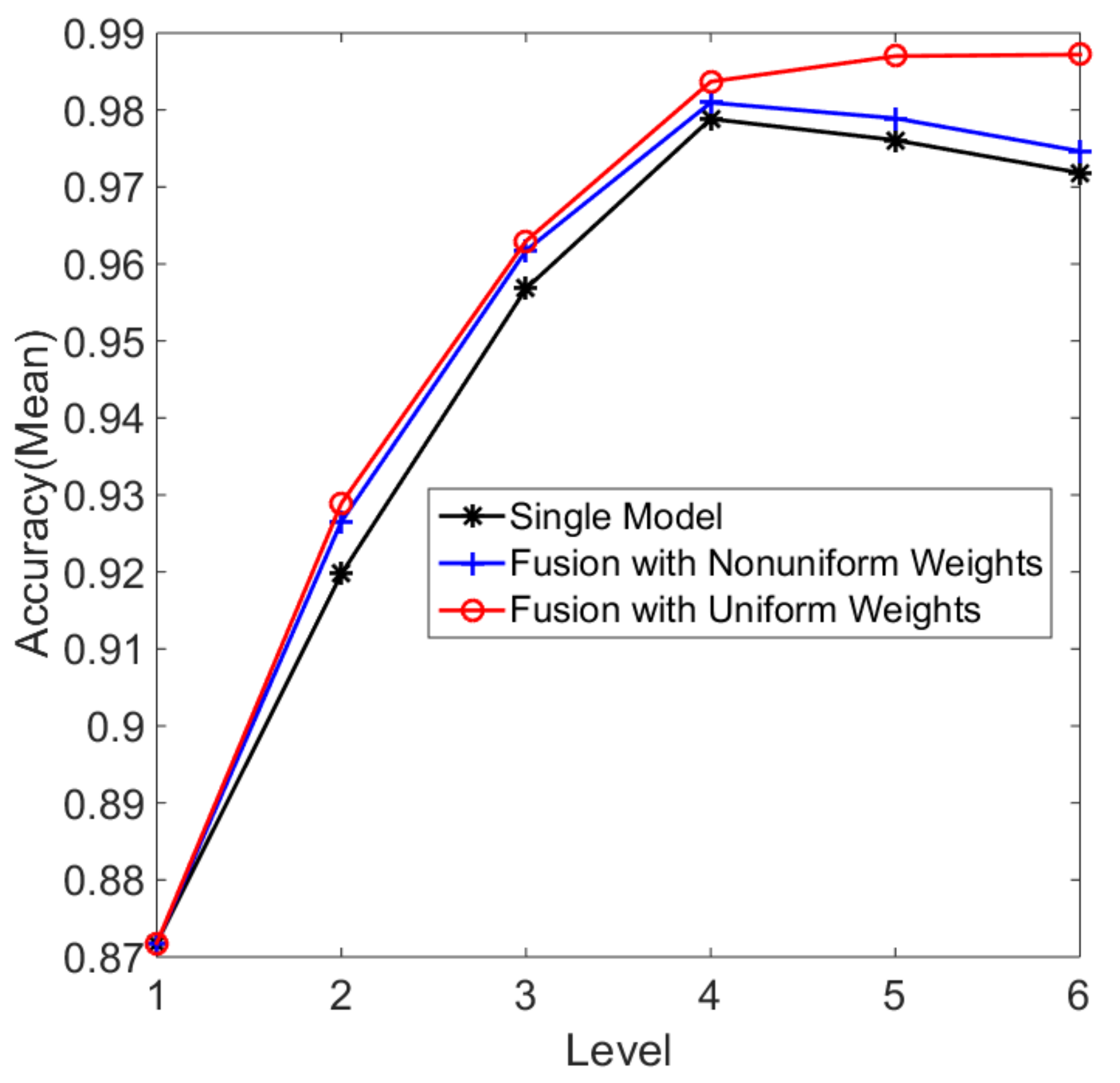}}
\hspace{0.025\linewidth}
\subfloat[]{\label{Fig:accurayMultiScaleWholeAugFusionStd}%%
\includegraphics[width=0.4\linewidth]{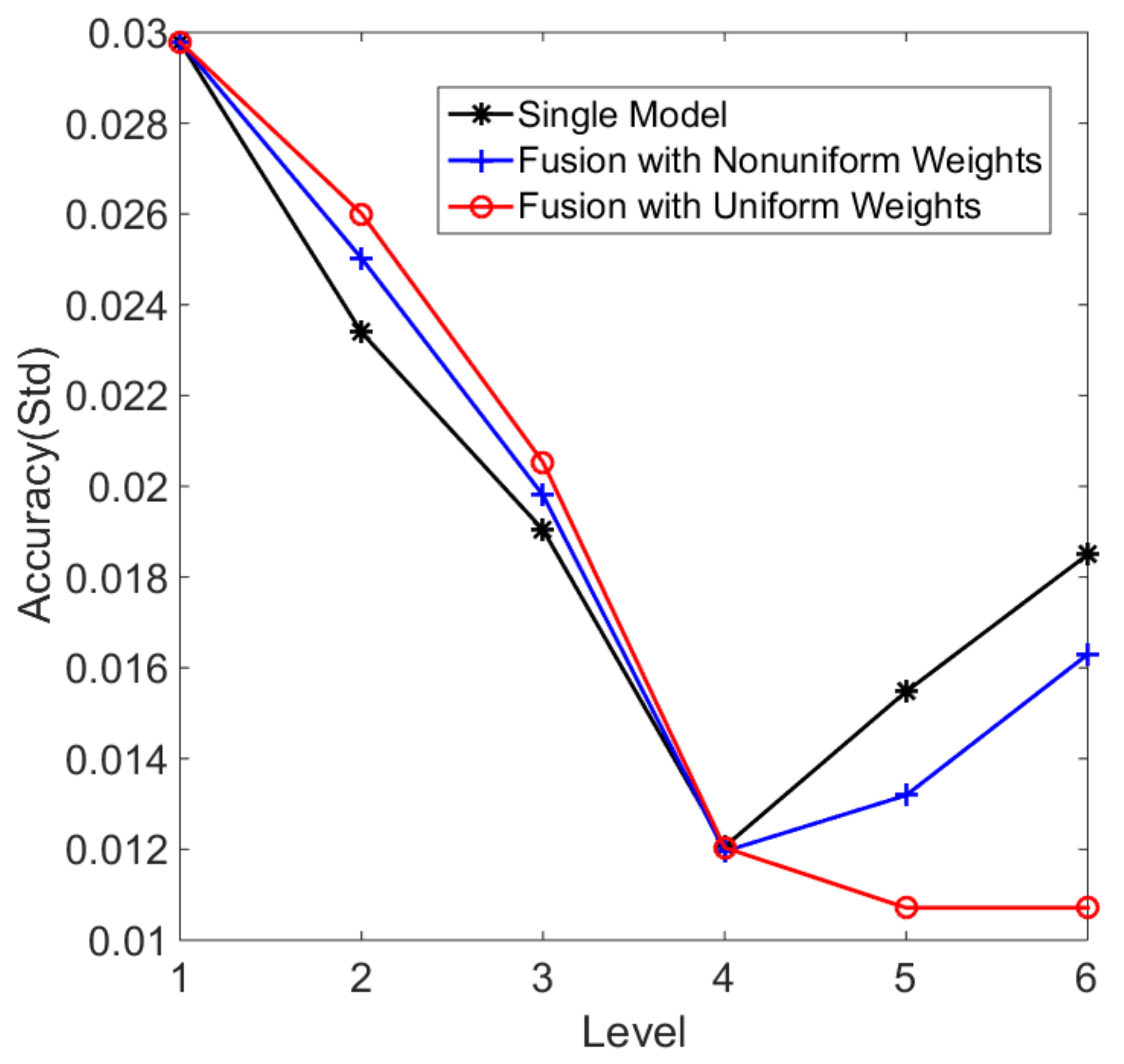}}

\caption{Results of the proposed online decision fusion method. \protect\subref{Fig:accurayMultiScaleWholeAugFusionMu} Means of the classification accuracy. \protect\subref{Fig:accurayMultiScaleWholeAugFusionStd} Standard deviations (Std) of the classification accuracy.}
\label{fig:accurayMultiScaleWholeAugFusion}
\end{figure*}

Besides, we also calculate the confusion matrix of the proposed method at the first level on the test set, which is shown in Figure~\ref{Fig:ConfusionMatrix}(a). It is clear that some categories are overlapped with others, $e.g.$, SA, VTr, AVB-I, and PVC. The results are consistent with the visual inspection results in Figure~\ref{fig:featureProj}.

\subsubsection{Online decision fusion performance}
\label{subsubsec:OnlineDecisionFusion}

We test the fusion method in Section~\ref{subsec:OnlineFusion} at different levels ${s_l}$: 2, 3, 4, 5, and 6. Two kinds of fusion weights are compared: the uniform one and the one favouring high level models which is calculated as:
\begin{equation}
{w_s} = \frac{{{2^{s - 1}}}}{{{2^{{s_l}}} - 1}}.
\label{eq:weight}
\end{equation}

Figure\ref{fig:accurayMultiScaleWholeAugFusion} shows the classification accuracy of the proposed fusion method and the proposed single scale method at different levels. First, it can be seen that $h4$ achieves the best performance among all the single models at different levels. The reason may be that it makes a trade-off between data length and the number of model decisions. Compared with $h1$, the input data length is 16 times larger. Compared with $h6$, which only makes a single decision on the whole sequence, $h4$ can make 4 decisions from different scale-specific models and fuse them into a more accurate one.

Then, it can be seen that the fusion results are consistently better than the results of the single-scale model. The performance is improved consistently with the growth of data length. It validates the idea that fusing the complementary decisions from different models leads to a more accurate and stable one. Besides, using non-uniform weights does not provide any advantage over the uniform one. The non-uniform weight strategy favors the higher-level models than the decisions from the lower-level models. Though it is better than the single model, the gains are indeed very marginal. Especially at higher levels, the performance is largely dominated by the model at the highest level. Please refer to Section~\ref{subsec:OnlineFusion} for more details. In conclusion, the proposed online decision fusion method with uniform weights at level 6 achieves the best result. For example, the accuracy is boosted from 87\% (single model at level 1) to 99\%, and the standard deviation is reduced from 0.03 (single model at level 1) to 0.011. Its sensitivity and specificity scores are shown in Table~\ref{tab:SensitivitySpecificityVal}, which shows a significant boost than other methods. The confusion matrix in Figure~\ref{Fig:ConfusionMatrix}(b) shows the similar results. These results demonstrate the effectiveness of the proposed online decision fusion method.

\begin{table}[htbp]
%\tiny
\newcommand{\tabincell}[2]{\begin{tabular}{@{}#1@{}}#2\end{tabular}}
  \renewcommand\arraystretch{0.8}
  \centering
  \caption{Running times (millisecond, ms) of the proposed method at different settings.}
    \begin{tabular}{ccccccc}
    \toprule
    Device & \multicolumn{2}{c}{Titan X}  & \multicolumn{2}{c}{TX2} & \multicolumn{2}{c}{TX2 (BatchSize: x10)} \\
    \midrule
    Level &GPU    &CPU    &GPU    &CPU    &GPU    &CPU \\
    \hline
    1 &0.01 &0.17    &0.17    &0.46    &0.12    &0.44 \\
    2 &0.03    &0.21    &0.27    &0.59    &0.13    &0.55 \\
    3 &0.05    &0.26    &0.37    &0.80    &0.14    &0.72 \\
    4 &0.10    &0.42    &1.22    &1.57    &0.18    &1.39 \\
    5 &0.21    &0.82    &2.01    &2.91    &0.27    &2.79 \\
    6 &0.33    &1.33    &2.73    &5.38    &0.56    &5.66 \\
    \bottomrule
    \end{tabular}%
  \label{tab:RunningTime}%
\end{table}%

\subsubsection{Computational complexity and running time analysis}
\label{subsubsec:ComputationalAnalysis}
We record the running times of the proposed method at GPU and CPU modes, respectively. Results are summarized in Table~\ref{tab:RunningTime}. As can be seen, the running time is only 0.33ms even if it is tested for the whole sequence (level 6). To further examine the computational efficiency of the proposed method, we test it on an NVIDIA Jetson TX2 embedded board. Again, the proposed method can achieve a real-time speed. Interestingly, the running times at GPU mode and CPU mode are comparable. We hypothesize that enlarging the batch size may make full advantage of the GPU acceleration. After enlarging the batch size 10 times, the superiority of GPU mode is significant. In conclusion, the proposed method is very efficient and promising to be integrated into a portable ECG monitor with limited computational resources.

\subsection{Experiments on a real-world ECG dataset}
\label{subsec:actualExperiments}

\begin{figure}
\centering
\includegraphics[width=0.8\linewidth]{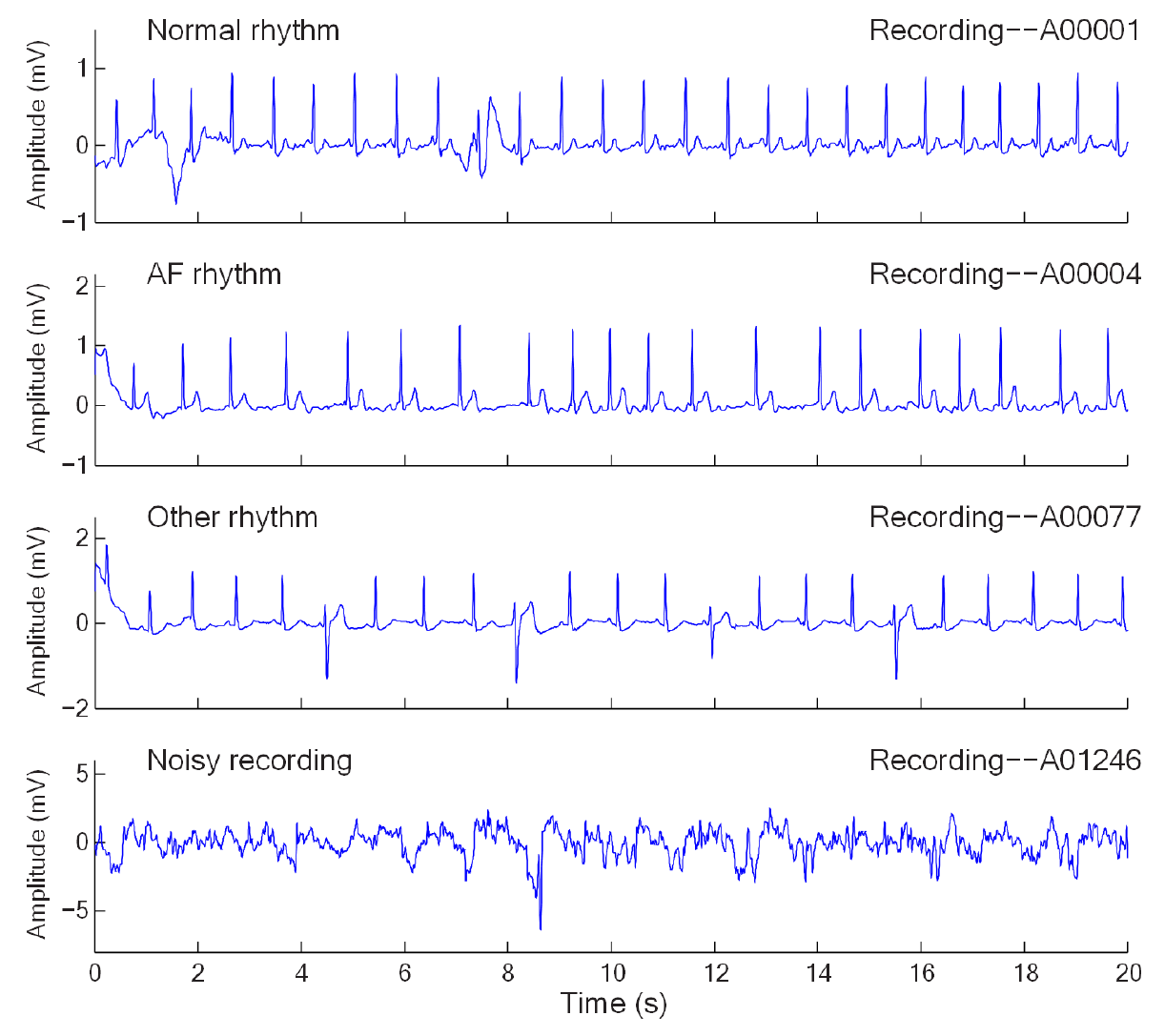}
\caption{Examples of the ECG waveforms in PhysioNet dataset \cite{clifford2017af}.}
\label{fig:physioNetExampleWaveforms}
\end{figure}

We also conducted extensive experiments on a real-world ECG dataset used in the 2017 PhysioNet/Computing in Cardiology Challenge \cite{clifford2017af}. The dataset is split into the training set, validation set, and test set. The training set contains 8,528 single-lead ECG recordings lasting from 9s to 60s. The validation set and test set contain 300 and 3,658 ECG recordings of similar lengths, respectively. The ECG recordings were sampled as 300 Hz. Each sample is labeled into four categories: Normal rhythm, AF rhythm, Other rhythm, and Noisy recordings. Only labels of the training set and validation set are publicly available. Some examples of the ECG waves are shown in Figure~\ref{fig:physioNetExampleWaveforms}.

We train our model on the training set. Scores both on the training set and validation set are reported and compared with the top entries in the challenge. It is noteworthy that we add two more convolutional layers after the first and second convolutional layers in the network depicted in Table~\ref{tab:Architecture} such that it has a stronger representation capacity to handle the real-world ECG signals better. The number of convolutional filters and kernel sizes are the same as their preceding counterparts. The first fully-connected layer is kept the same. The output number of the last fully-connected layer is modified to be four to keep consistent with the number of categories. Each sample in the dataset is cropped or duplicated to have a length of 16,384. All other hyper-parameters are kept the same as the above experiments if not specified. We train the model at each level three times with random seeds and report the average scores and standard deviations.

\subsubsection{Comparisons of the proposed method and the top entries in the challenge}
\label{subsubsec:onlineFusionPhysioNet}

\begin{figure*}
\centering
\subfloat[]{\label{Fig:onlineFusionAccMean_PhysioNet}%%
\includegraphics[width=0.4\linewidth]{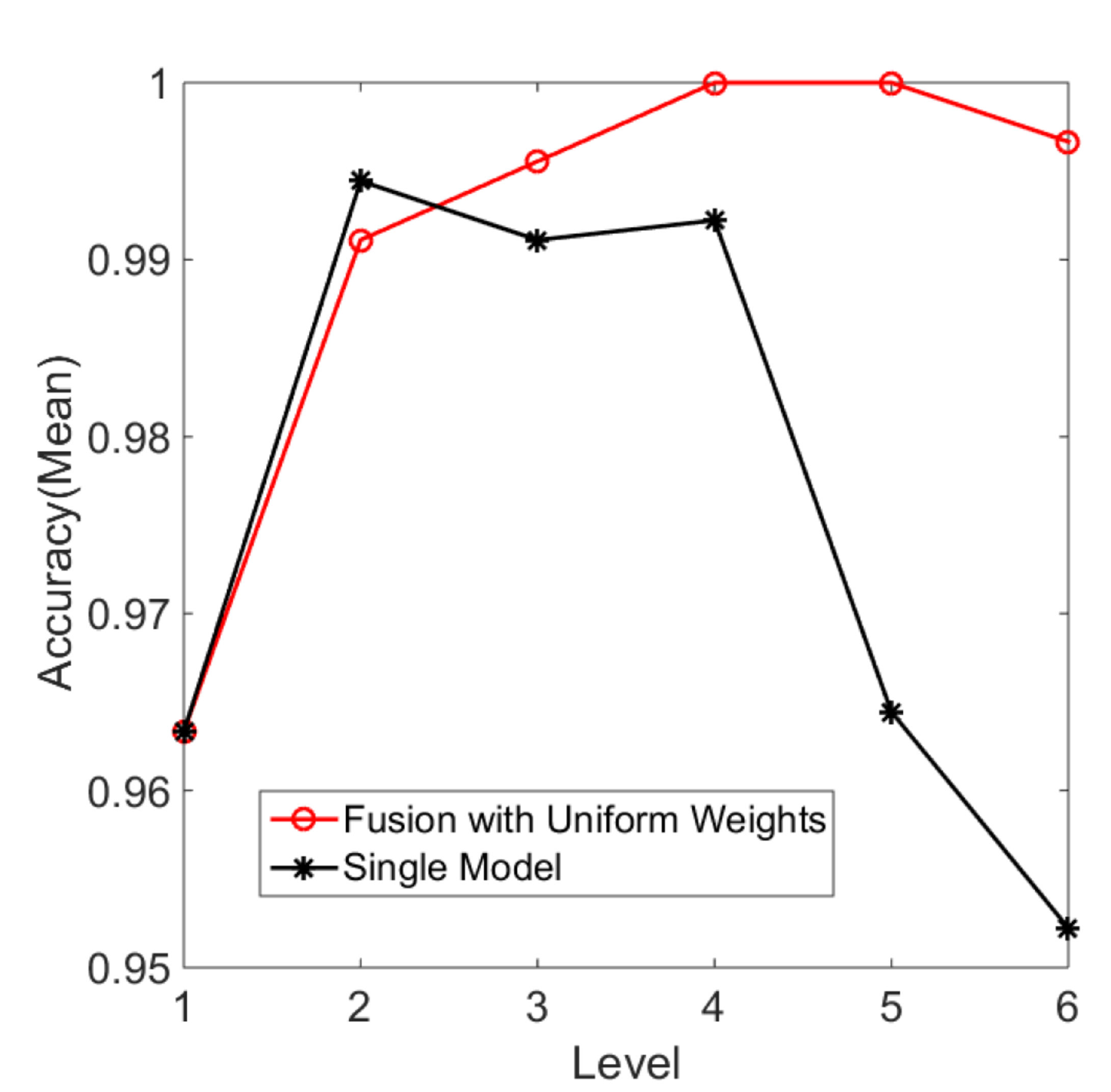}}
\hspace{0.025\linewidth}
\subfloat[]{\label{Fig:onlineFusionAccStd_PhysioNet}%%
\includegraphics[width=0.38\linewidth]{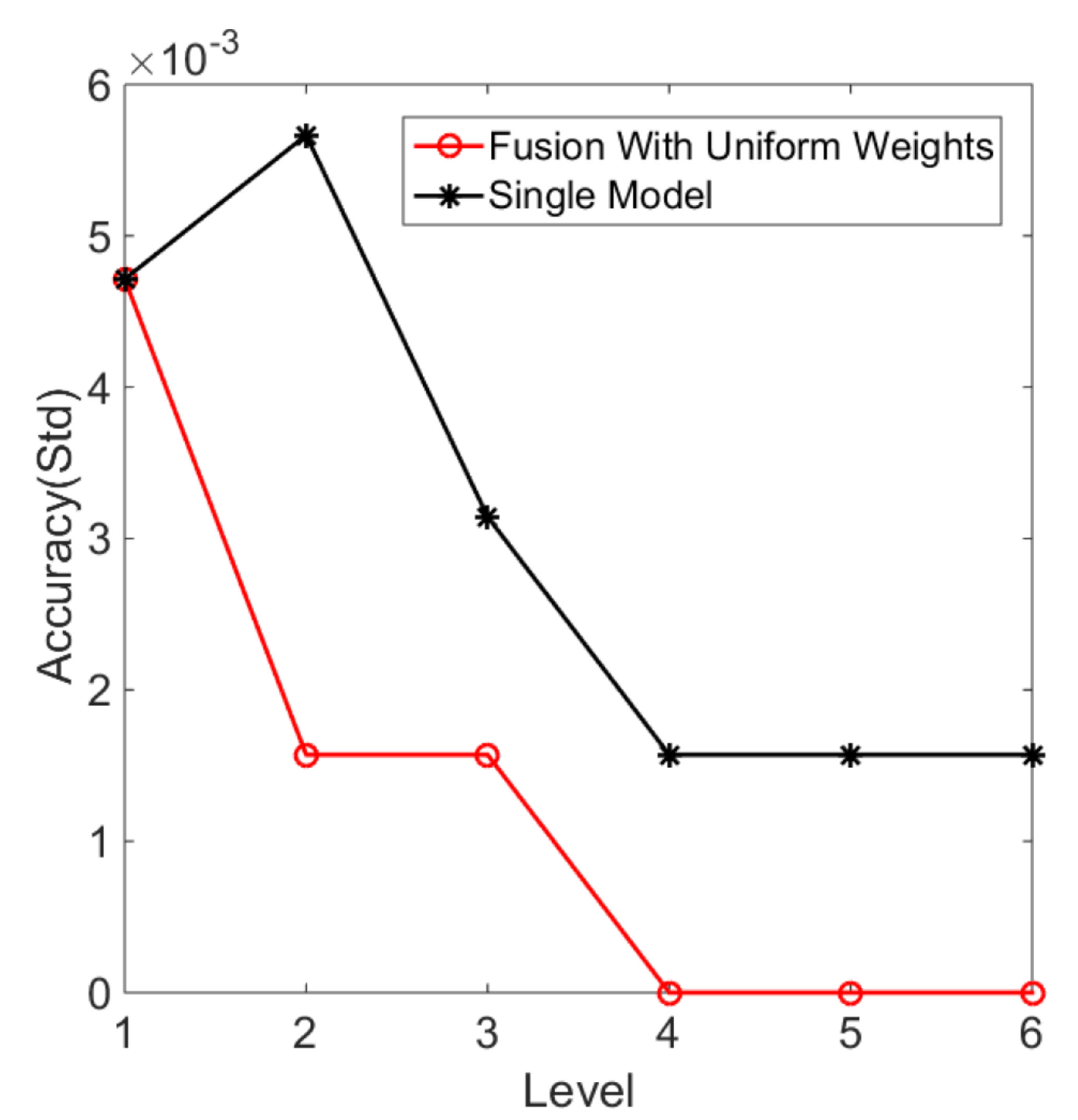}}

\caption{Results of the proposed online decision fusion method on the PhysioNet dataset \cite{clifford2017af}. \protect\subref{Fig:onlineFusionAccMean_PhysioNet} Mean values of the classification accuracy. \protect\subref{Fig:onlineFusionAccStd_PhysioNet} Standard deviations (Std) of the classification accuracy.}
\label{fig:onlineFusionAcc_PhysioNet}
\end{figure*}

\begin{figure*}
\centering
\subfloat[]{\label{Fig:onlineFusionF1Mean_PhysioNet}%%
\includegraphics[width=0.4\linewidth]{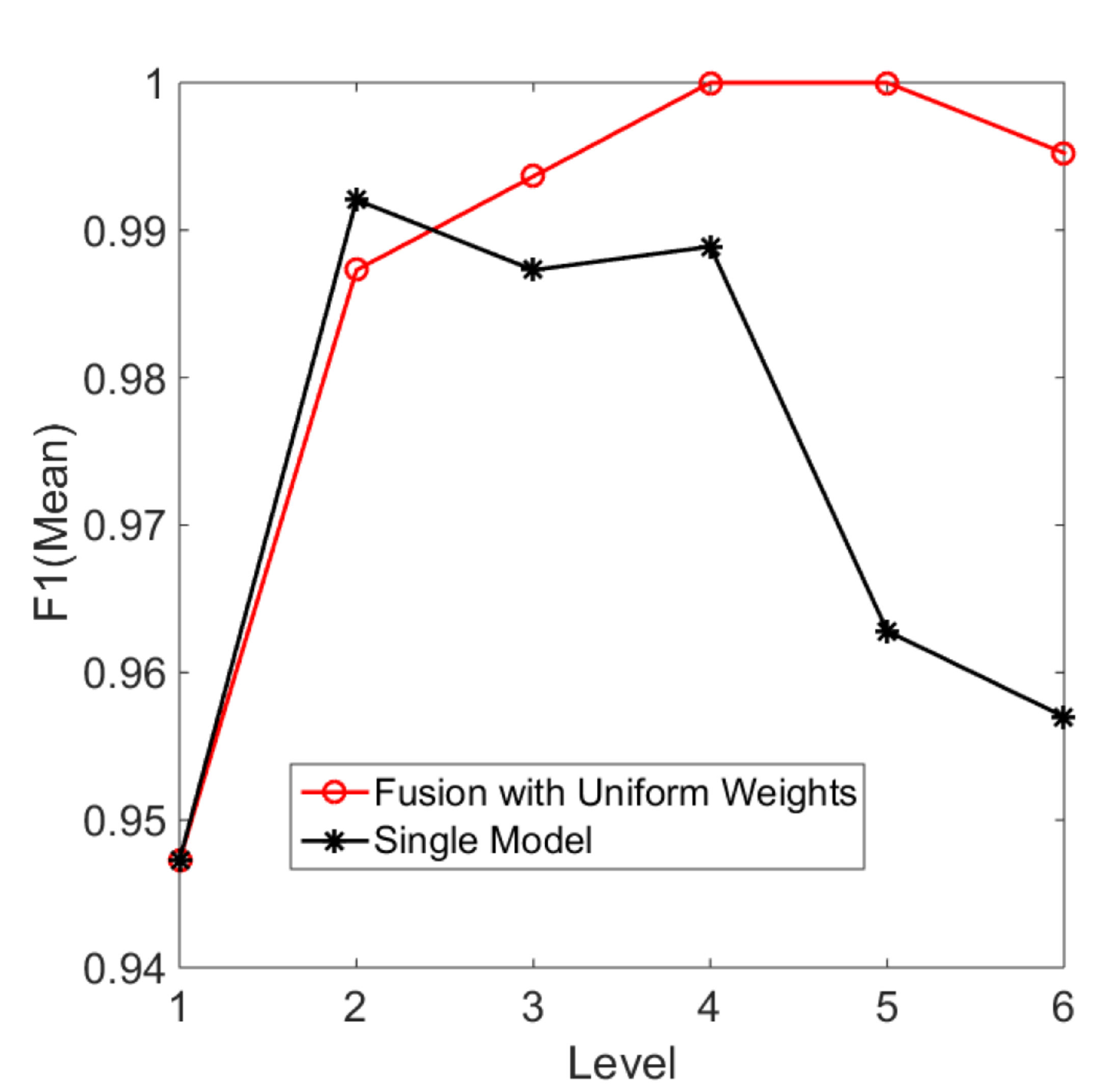}}
\hspace{0.025\linewidth}
\subfloat[]{\label{Fig:onlineFusionF1Std_PhysioNet}%%
\includegraphics[width=0.38\linewidth]{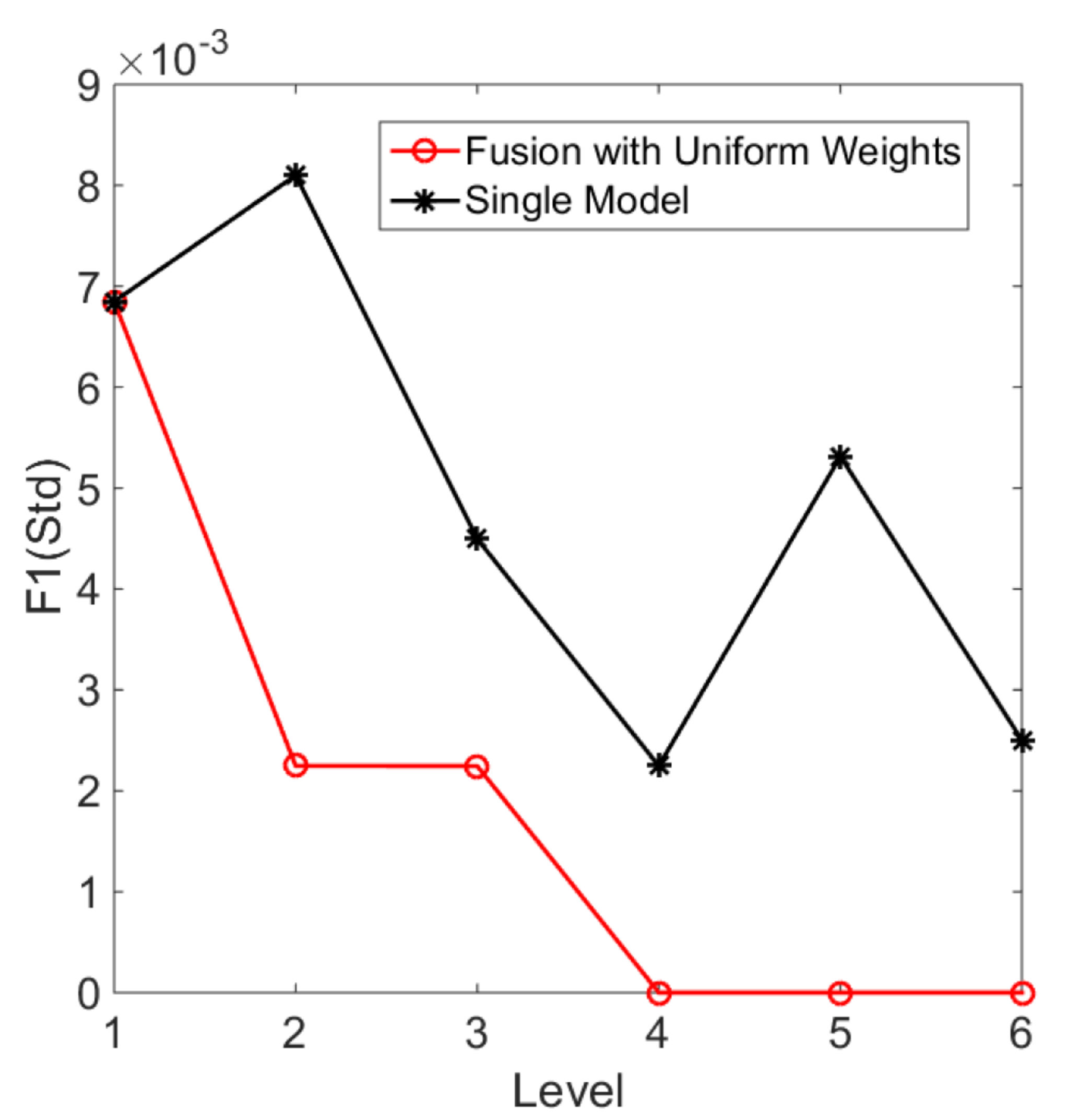}}

\caption{Results of the proposed online decision fusion method on the PhysioNet dataset \cite{clifford2017af}. \protect\subref{Fig:onlineFusionF1Mean_PhysioNet} Mean values of F1 scores. \protect\subref{Fig:onlineFusionF1Std_PhysioNet} Standard deviations (Std) of the F1 scores.}
\label{fig:onlineFusionF1_PhysioNet}
\end{figure*}

\begin{figure}
\centering
\includegraphics[width=0.5\linewidth]{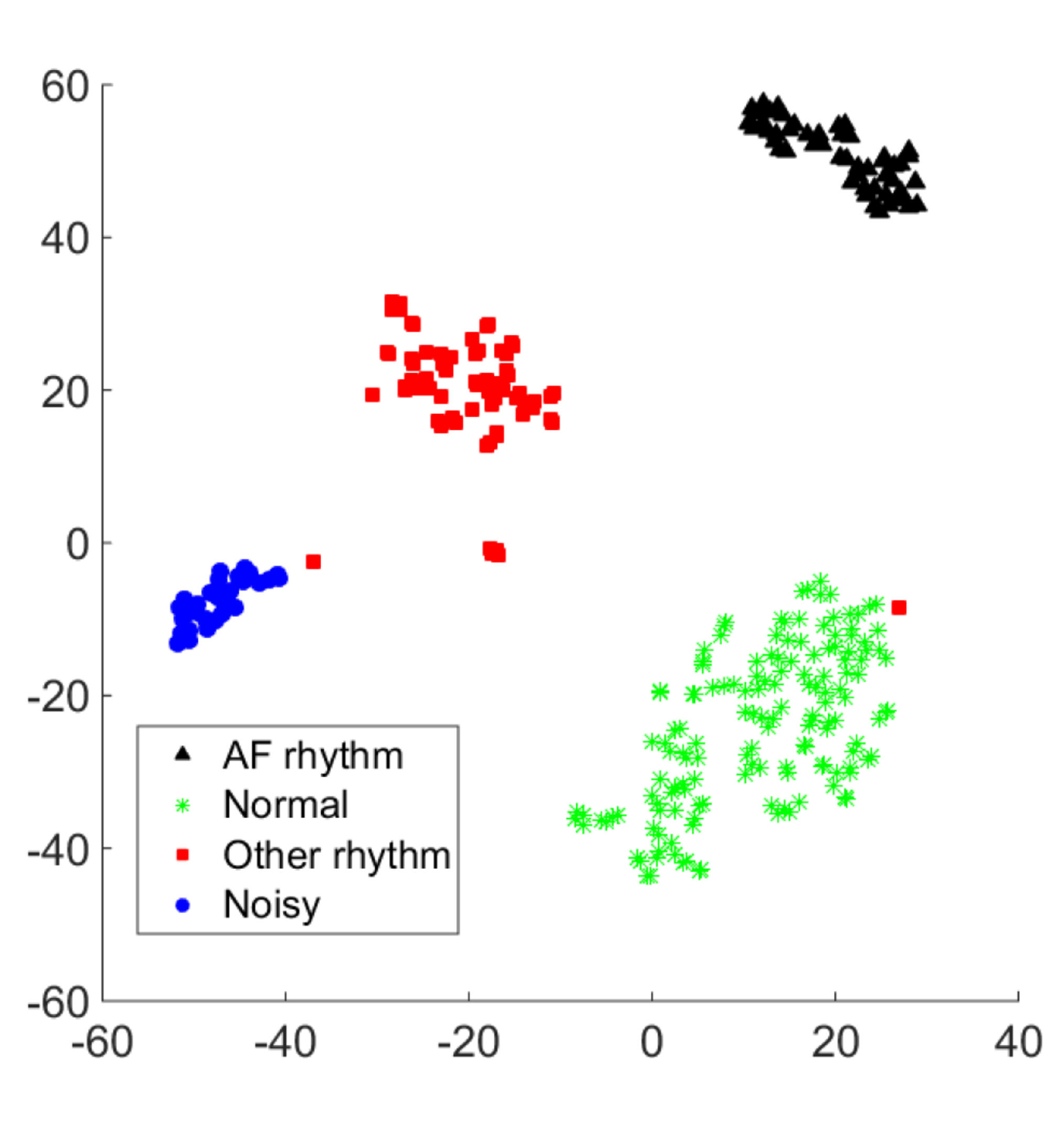}
\caption{Visualization of the learned features from the proposed network on the PhysioNet dataset \cite{clifford2017af}.}
\label{fig:featureProj_PhysioNet}
\end{figure}

We report the results on this dataset in terms of mean accuracy and F1 score. To keep consistent with the evaluation protocol in \cite{clifford2017af}, we report the average F1 scores for the first three categories. Besides, we also include the average F1 scores for all categories. As can be seen from Figure~\ref{fig:onlineFusionAcc_PhysioNet} and Figure~\ref{fig:onlineFusionF1_PhysioNet}, the best results are achieved at level 4 ($h4$) and level 5 ($h5$) by the proposed online fusion method, $i.e.$, $100\% \pm 0\%$ classification accuracy and $100\% \pm 0\%$ F1 score. Meanwhile, the results of single models are also competitive. Best results are achieved at level 2 ($h2$) with a $99.44\% \pm 0.57\%$ accuracy and a $99.21\% \pm 0.81\%$ F1 score. It is consistent with the result in Section~\ref{subsubsec:OnlineDecisionFusion}, where model $h4$ makes a trade-off between data length and the number of model decisions. The comparison results between the proposed method and the top entries in the challenge are listed in Table~\ref{tab:Challenge-2017}. The proposed methods achieve comparable or better results than the top entries on both the training set and validation set.

% Table generated by Excel2LaTeX from sheet 'Sheet1'
\begin{table*}[htbp]
  \centering
  \caption{Accuracy and F1 scores on the PhysioNet dataset of the proposed methods and the top entries in the challenge. ``F1 score'' stands for the average F1 score for the first three categories, $i.e.$, Normal rhythm, AF rhythm, and Other rhythm. ``F1 score (all categories)'' stands for the average F1 score for all categories. }
    \resizebox{\textwidth}{30mm}{
    \begin{tabular}{cccccccc}
    \toprule
    Rank & Entry & \multicolumn{2}{c}{accuracy} & \multicolumn{2}{c}{F1 score} & \multicolumn{2}{c}{F1 score (all categories)} \\
    \multicolumn{1}{c}{} & \multicolumn{1}{c}{} & Validation & Train & Validation & Train & Validation & Train \\
    \midrule
    1     & \textit{Teijeiro $et~al.$} \cite{teijeiro2017arrhythmia} & -     & -     & 0.912 & 0.893 & -     & - \\
    1     & \textit{Datta $et~al.$} & -     & -     & 0.990  & 0.970  & -     & - \\
    1     & \textit{Zabihi $et~al.$} \cite{zabihidetection} & -     & -     & 0.968 & 0.951 & -     & - \\
    1     & \textit{Hong $et~al.$} \cite{hong2017encase} & -     & -     & 0.990  & 0.970  & -     & - \\
    5     & \textit{Baydoun $et~al.$} & -     & -     & 0.859 & 0.965 & -     & - \\
    5     & \textit{Bin $et~al.$} & -     & -     & 0.870  & 0.875 & -     & - \\
    5     & \textit{Zihlmann $et~al.$} & -     & -     & 0.913 & 0.889 & -     & - \\
    5     & \textit{Xiong $et~al.$} & -     & -     & 0.905 & 0.877 & -     & - \\
    \hline
    -     & \textit{Proposed (level 4)} & 0.992$\pm$0.002 & 0.998$\pm$0.001 & 0.989$\pm$0.002 & \textbf{0.996$\pm$0.002} & 0.992$\pm$0.002 & 0.995$\pm$0.003 \\
    -     & \textit{Proposed (fusion, level 4)} & 1.0$\pm$0.0 & 0.999$\pm$0.001 & \textbf{1.0$\pm$0.0}  & 0.994$\pm$0.006 & 1.0$\pm$0.0  & 0.991$\pm$0.009 \\
    \bottomrule
    \end{tabular}}%
  \label{tab:Challenge-2017}%
\end{table*}%

\subsubsection{Analysis on learned features}
\label{subsubsec:AnalysisFeatures_PhysioNet}
Similar to Section~\ref{subsubsec:AnalysisFeatures}, we plot the learned features from model $h4$ on the validation set in Figure~\ref{fig:featureProj_PhysioNet}. As can be seen, samples in each category are almost clustered together and separated from other clusters. For several samples in the category of ''Other rhythm'', they are near the clusters of ''Normal'' and ''Noisy''. It implies that these samples are either with noise labels or hard cases which should be carefully handled.

\begin{figure*}
\centering
\subfloat[]{\label{Fig:ss-AT}%%
\includegraphics[width=0.4\linewidth]{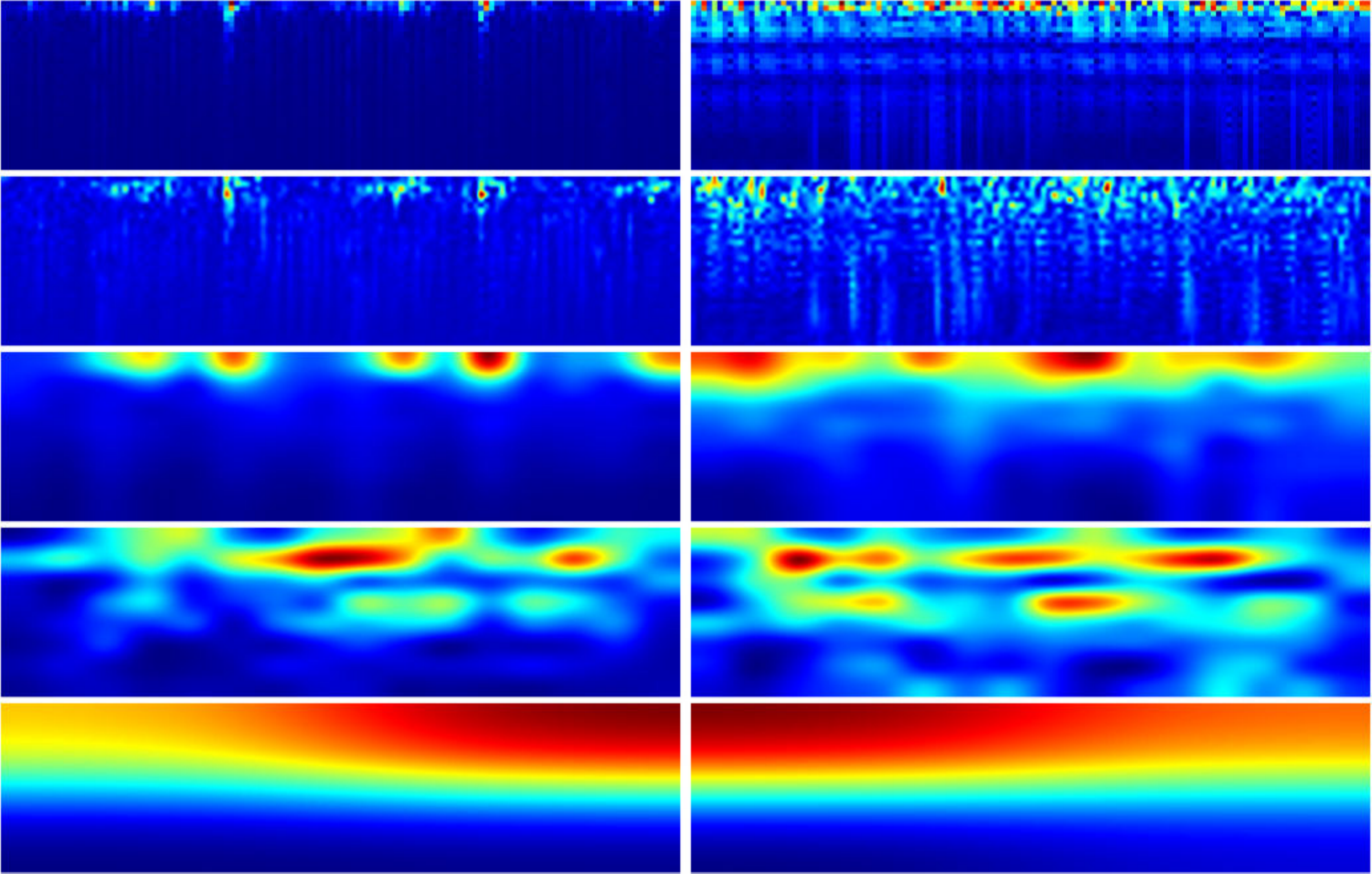}}
\hspace{0.025\linewidth}
\subfloat[]{\label{Fig:ss-Normal}%%
\includegraphics[width=0.4\linewidth]{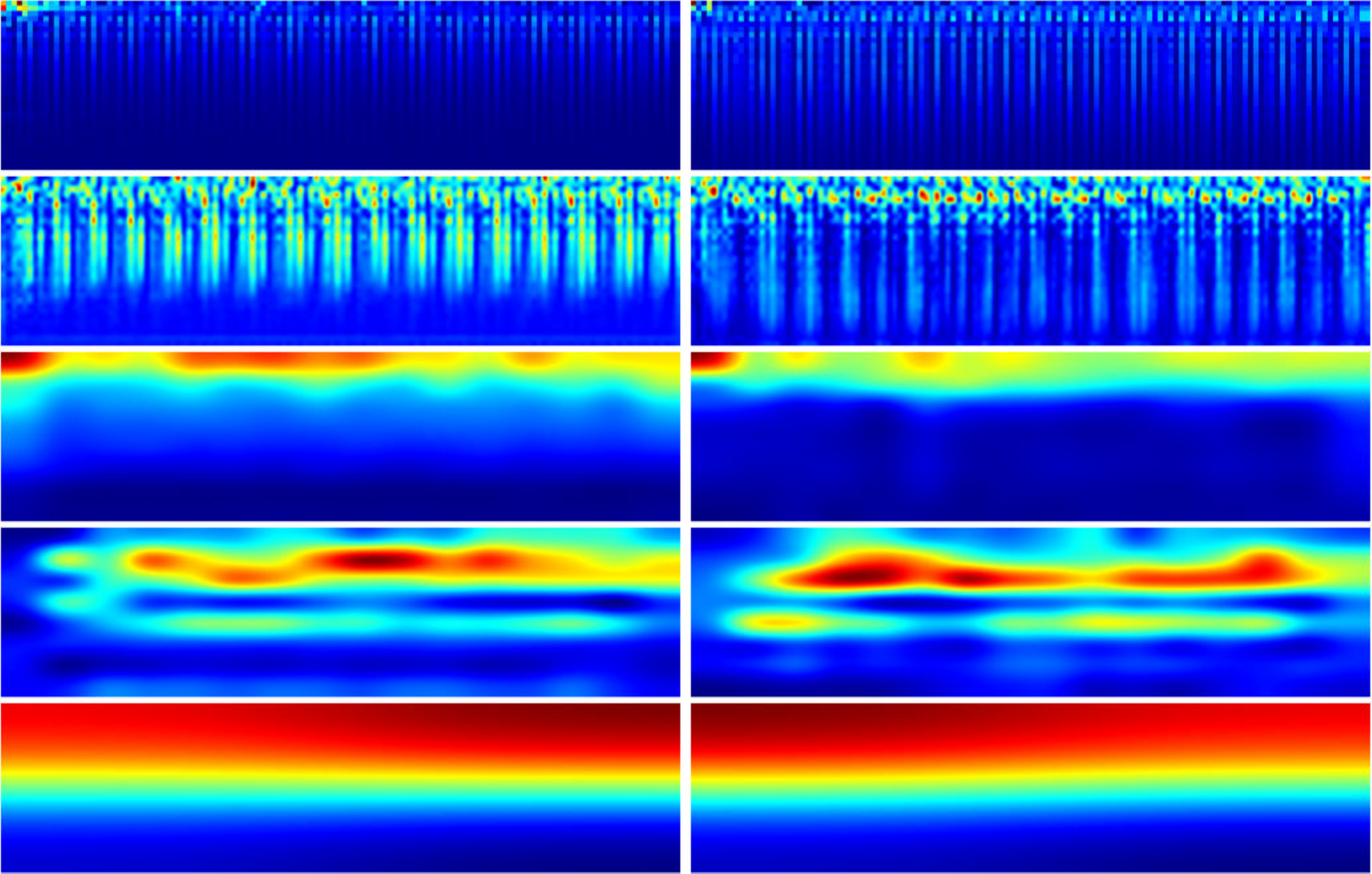}}

\subfloat[]{\label{Fig:ss-other}%%
\includegraphics[width=0.4\linewidth]{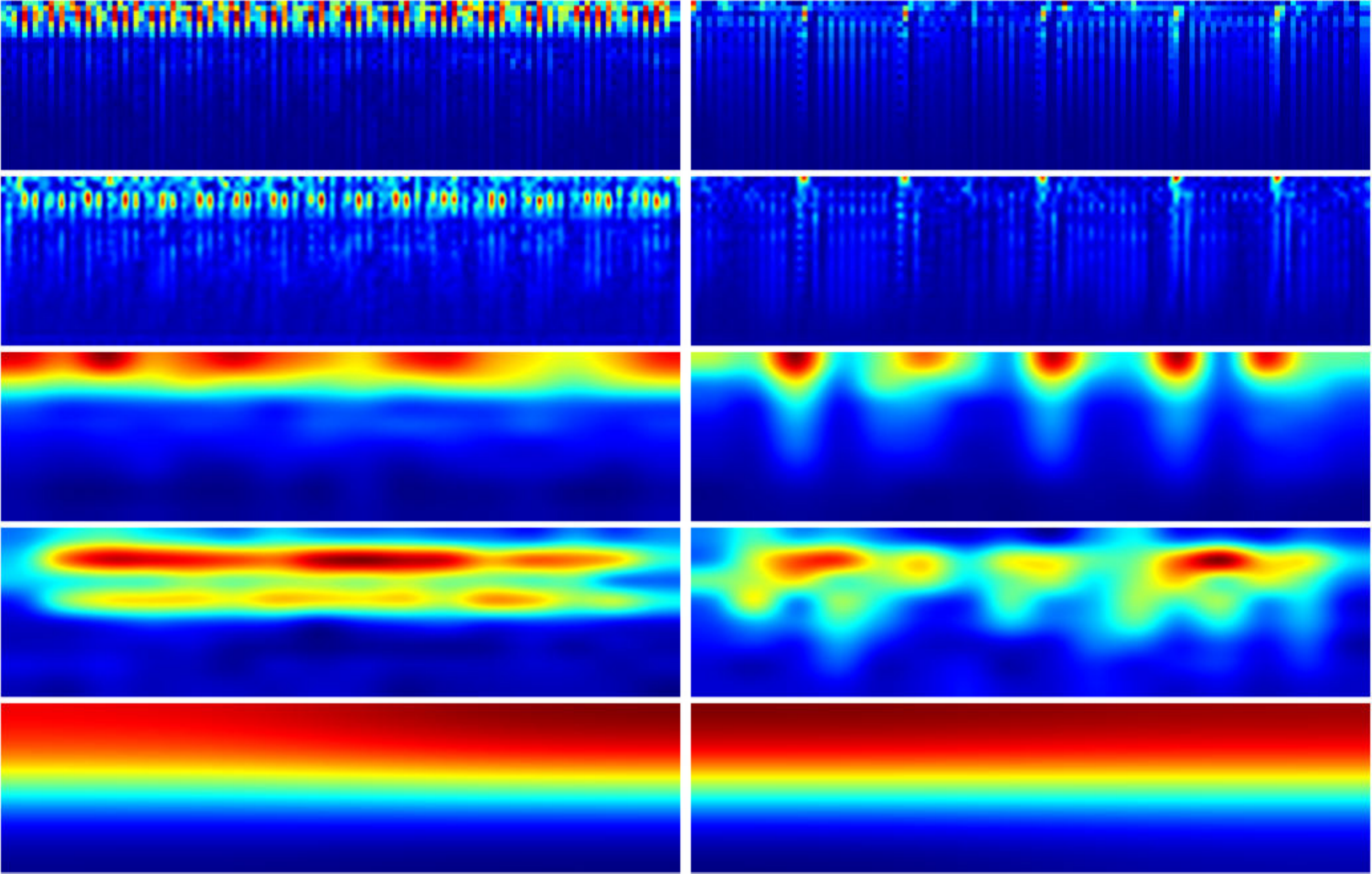}}
\hspace{0.025\linewidth}
\subfloat[]{\label{Fig:ss-noise}%%
\includegraphics[width=0.4\linewidth]{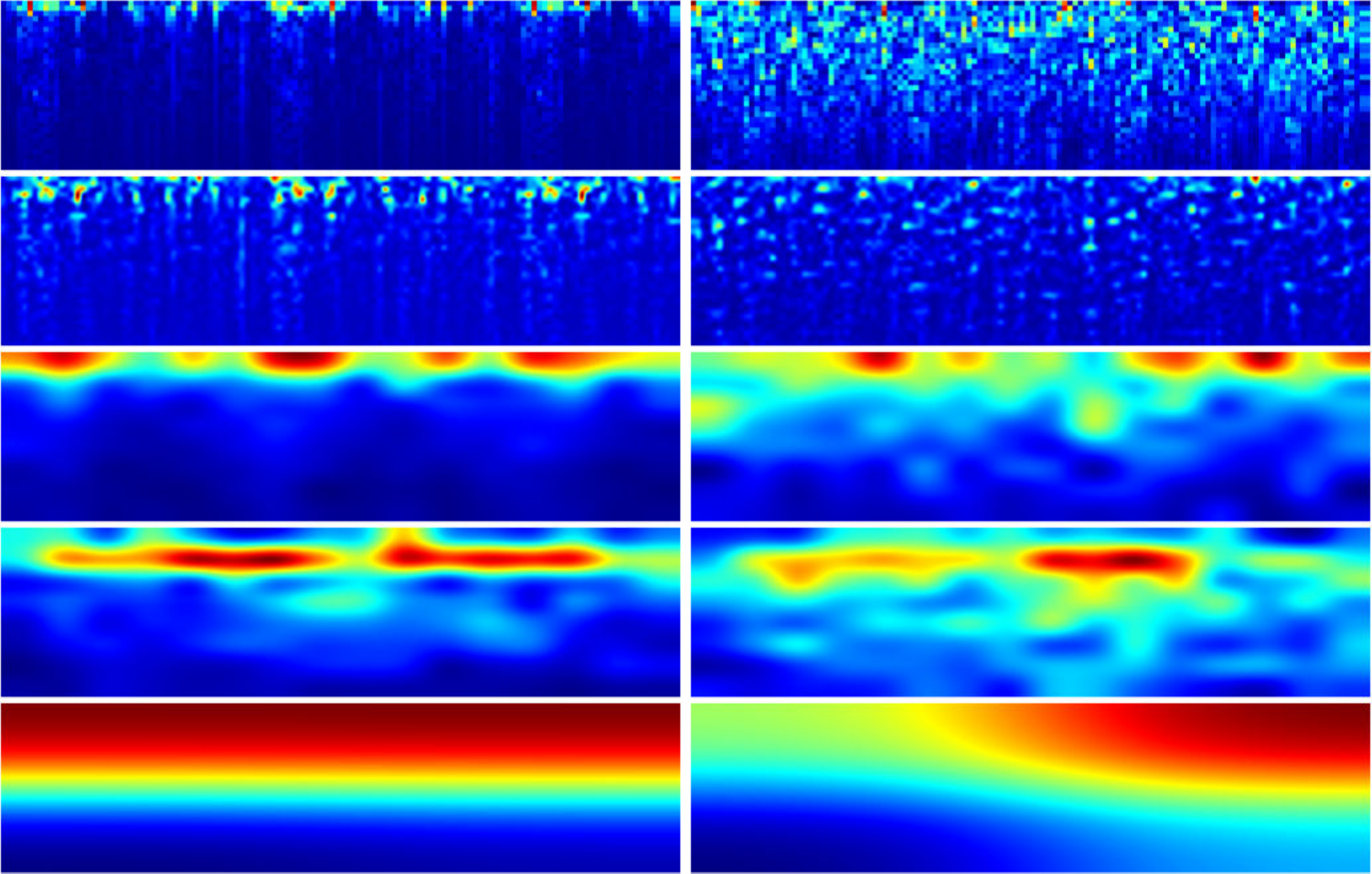}}

\caption{Spectrograms and corresponding feature maps. (a)-(d) shows the results of two samples from AF rhythm, Normal rhythm, Other rhythm, and Noisy rhythm, respectively. In each figure, the spectrograms and corresponding feature maps from $Conv1$, $Pool1$, $Conv2$ and $Pool2$ layers are plotted from the top row to the bottom row, respectively. Hot color represents a strong response.}
\label{fig:featResponse}
\end{figure*}

In addition, we plot the spectrograms and their corresponding feature maps from $Conv1$, $Pool1$, $Conv2$ and $Pool2$ layers in Figure~\ref{fig:featResponse}. As can be seen, the first convolutional layer acts like a basic feature extractor which strengthens the informative parts in the spectrograms. Then, features in low and medium frequencies are pooled and contribute to the final classification. From the $Conv2$ feature maps, we can see that the proposed network generates strong responses in specific frequency zones and accumulate them along the temporal axis. By doing so and together with the online fusion, it learns effective and discriminative features to make an accurate classification.

%%%%%%%%%%%%%%%%%%%%%%%%%%%%%%%%%%%%%%%%%%
\section{Conclusion and future work}
\label{sec:conclusion}
In this paper, we propose a novel deep CNN based method for ECG signal classification. It learns discriminative feature representation from the time-frequency domain by calculating the Short-Time Fourier Transform of the original wave signal. Besides, the proposed online decision fusion method fuses complementary decisions from different scale-specific models into a more accurate one. Extensive experiments on a synthetic 20-category ECG dataset and a real-world AF classification dataset demonstrate its effectiveness. Moreover, the proposed method is computationally efficient and promising to be integrated into a portable ECG monitor with limited computational resources. Future research may include: 1) devising or searching compact and efficient networks to handle complex real-world ECG data; 2) improving the online fusion by integrating both the decisions and learned features at different levels; 3) exploring the potential of the proposed method for nonlinear time series beyond ECG.

\section*{Acknowledgment}
This work was partly supported by the National Natural Science Foundation of China (NSFC) under Grants 61806062, 61873077, and 61872327, the NSFC-Zhejiang Joint Fund for the Integration of Industrialization and Informatization under the Grant U1709215, the Fundamental Research Funds for the Central Universities under Grant WK2380000001, and the Zhejiang Province Key R\&D Project under Grant 2019C03104.

%\section*{References}

\bibliography{KBS_ECG}

\end{document}